\documentclass[lettersize,journal]{IEEEtran}
\usepackage{times}
\usepackage{epsfig}
\usepackage{graphicx}
\usepackage{amsmath}
\usepackage{amssymb}
\usepackage{booktabs}
\usepackage{algorithm}
\usepackage{algpseudocode}
\usepackage[table,xcdraw,dvipsnames]{xcolor}
\usepackage{multirow}
\usepackage{subcaption}
\usepackage{enumitem}
\usepackage{balance}
\usepackage{mathrsfs} 
\usepackage{bm}
\usepackage{wrapfig}
\usepackage{bbm}
\usepackage{stmaryrd}

\makeatletter
\newcommand*\bigcdot{\mathpalette\bigcdot@{.5}}
\newcommand*\bigcdot@[2]{\mathbin{\vcenter{\hbox{\scalebox{#2}{$\m@th#1\bullet$}}}}}
\makeatother

\definecolor{c2}{HTML}{FBD9BD}
\definecolor{c3}{HTML}{fe793d}
\definecolor{c4}{HTML}{eedeb0}
\definecolor{c5}{HTML}{00FFFF}
\definecolor{c6}{HTML}{FF00FF}

\definecolor{rouse}{rgb}{0.981,0.961,0.941}

\usepackage[pagebackref=true,breaklinks=true,letterpaper=true,colorlinks,bookmarks=false]{hyperref}

\usepackage[capitalize]{cleveref}
\crefname{section}{Sec.}{Secs.}
\Crefname{section}{Section}{Sections}
\Crefname{table}{Table}{Tables}

\ifCLASSOPTIONcompsoc
  \usepackage[nocompress]{cite}
\else
  \usepackage{cite}
\fi

\ifCLASSINFOpdf
\else
\fi

\begin{document}
%

\title{
Reversible Unfolding Network for Concealed \\ Visual Perception with Generative Refinement}

\author{
Chunming~He,~\IEEEmembership{}
Fengyang Xiao,
Rihan~Zhang, 
Chengyu Fang, \\
Deng-Ping Fan,~\IEEEmembership{Senior Member,~IEEE,} 
and Sina Farsiu,~\IEEEmembership{Fellow,~IEEE.}
\IEEEcompsocitemizethanks{
\IEEEcompsocthanksitem This research was supported by the Foundation Fighting Blindness (BR-CL-0621-0812-DUKE); Research to Prevent Blindness (Unrestricted Grant to Duke University), Foundation Fighting Blindness (PPA-1224-0890-DUKE), and NSFC (No. 62476143). (\textit{Corresponding author:  
Sina Farsiu (E-mail: sina.farsiu@duke.edu)} and Deng-Ping Fan (E-mail: fdp@nankai.edu.cn))
\IEEEcompsocthanksitem Chunming He, Fengyang Xiao, Rihan Zhang, and Sina Farsiu are with the Department of Biomedical Engineering, Duke University, Durham, NC 27708 USA (e-mail: chunming.he@duke.edu, sina.farsiu@duke.edu).
\IEEEcompsocthanksitem Chengyu Fang is with Tsinghua Shenzhen International Graduate School, Tsinghua University, Shenzhen 518055, China. 
\IEEEcompsocthanksitem Deng-Ping Fan is with Nankai International Advanced Research Institute (SHENZHEN FUTIAN), Nankai University, Shenzhen, 518045, China, and he is also with College of Computer Science, Nankai University, Tianjin, 300071, China (E-mail: fdp@nankai.edu.cn).
}
}

\markboth{Submitted to IEEE TPAMI}%
{He \MakeLowercase{\textit{et al.}}: RUN++: Reversible Unfolding Network for Concealed Visual Perception with Generative Refinement}

\maketitle
\begin{abstract}
Existing methods for concealed visual perception (CVP) often leverage reversible strategies to decrease uncertainty, yet these are typically confined to the mask domain, leaving the potential of the RGB domain underexplored. 
To address this, we propose a reversible unfolding network with generative refinement, termed RUN++. 
Specifically, RUN++ first formulates the CVP task as a mathematical optimization problem and unfolds the iterative solution into a multi-stage deep network. This approach provides a principled way to apply reversible modeling across both mask and RGB domains while leveraging a diffusion model to resolve the resulting uncertainty.
Each stage of the network integrates three purpose-driven modules: a Concealed Object Region Extraction (CORE) module applies reversible modeling to the mask domain to identify core object regions; a Context-Aware Region Enhancement (CARE) module extends this principle to the RGB domain to foster better foreground-background separation; and a Finetuning Iteration via Noise-based Enhancement (FINE) module provides a final refinement. 
The FINE module introduces a targeted Bernoulli diffusion model that refines only the uncertain regions of the segmentation mask, harnessing the generative power of diffusion for fine-detail restoration without the prohibitive computational cost of a full-image process.
This unique synergy, where the unfolding network provides a strong uncertainty prior for the diffusion model, allows RUN++ to efficiently direct its focus toward ambiguous areas, significantly mitigating false positives and negatives. 
Furthermore, we introduce a new paradigm for building robust CVP systems that remain effective under real-world degradations and extend this concept into a broader bi-level optimization framework. This establishes a unified strategy that synergistically combines low-level and high-level vision tasks, enhancing the robustness of high-level applications.
Extensive experiments on a wide array of tasks—including standard, label-deficient, multimodal, and video CVP—demonstrate the state-of-the-art performance and flexibility of RUN++, establishing a new benchmark for the field. We will release the code and models.

\end{abstract}



\IEEEpeerreviewmaketitle

\setlength{\abovedisplayskip}{2pt}
\setlength{\belowdisplayskip}{2pt}

\section{Introduction}\label{sec:introduction}
\begin{figure}[!h]
\setlength{\abovecaptionskip}{0cm}
	\centering
	\includegraphics[width=\linewidth]{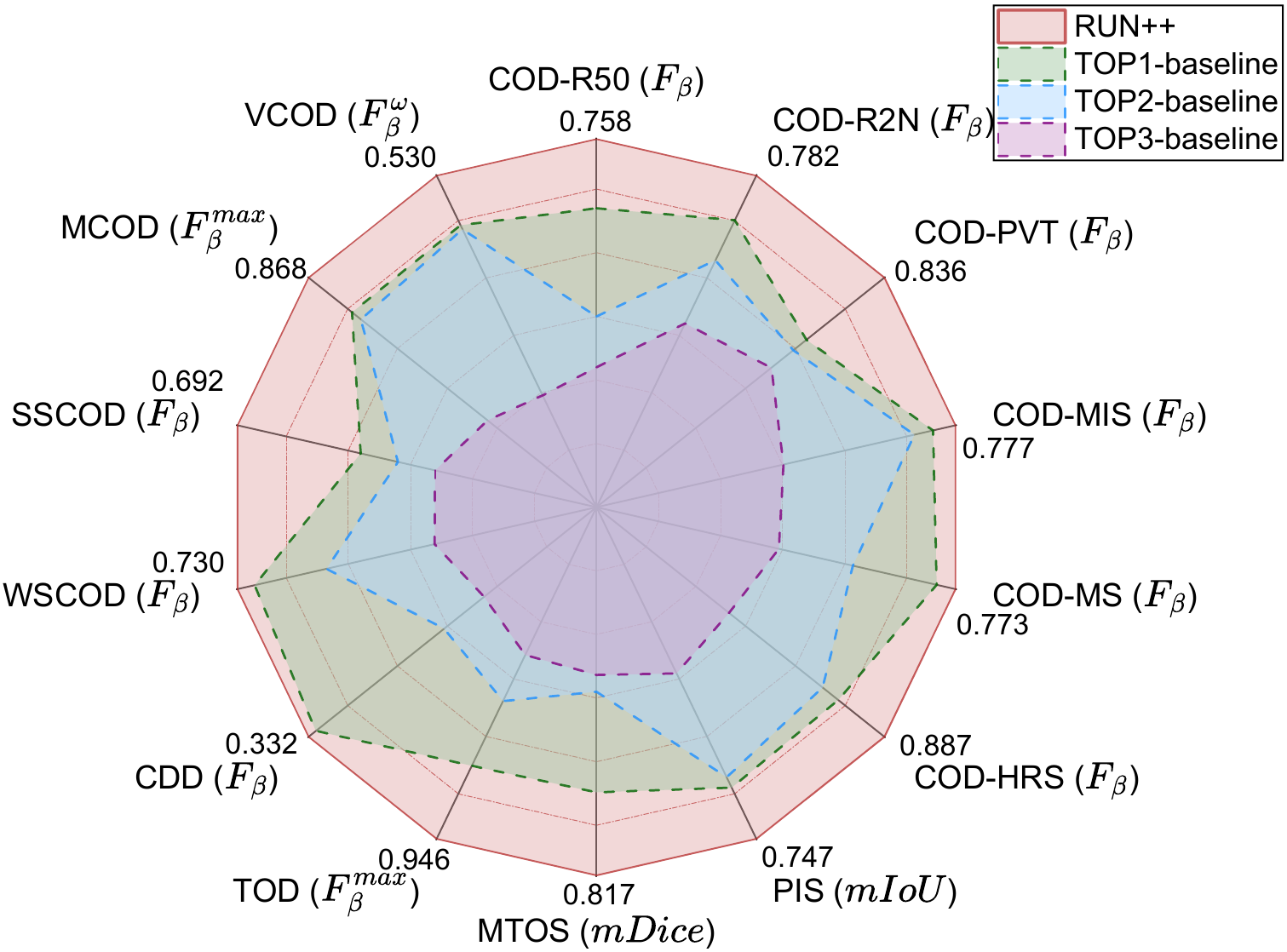}
	\caption{Comparisons between our RUN++ with cutting-edge methods in 14 CVP tasks using different datasets, backbones, and settings. The TOP1-Baseline is a composite benchmark, where the best-performing SOTA model is selected for each task. We evaluate performance on \textit{COD10K} (COD-R50, COD-R2N, COD-PVT, COD-MIS, COD-MS, COD-HRS, WSCOD, SSCOD, and MCOD), \textit{CVC} (PIS), \textit{DRIVE} (MTOS), \textit{GSD} (TOD), \textit{CDS2k} (CDD), and \textit{MoCA} (VCOD).
    }
	\label{fig:Radar}
	\vspace{-4mm}
\end{figure}
\begin{figure}[!h]
\setlength{\abovecaptionskip}{0cm}
	\centering
	\includegraphics[width=\linewidth]{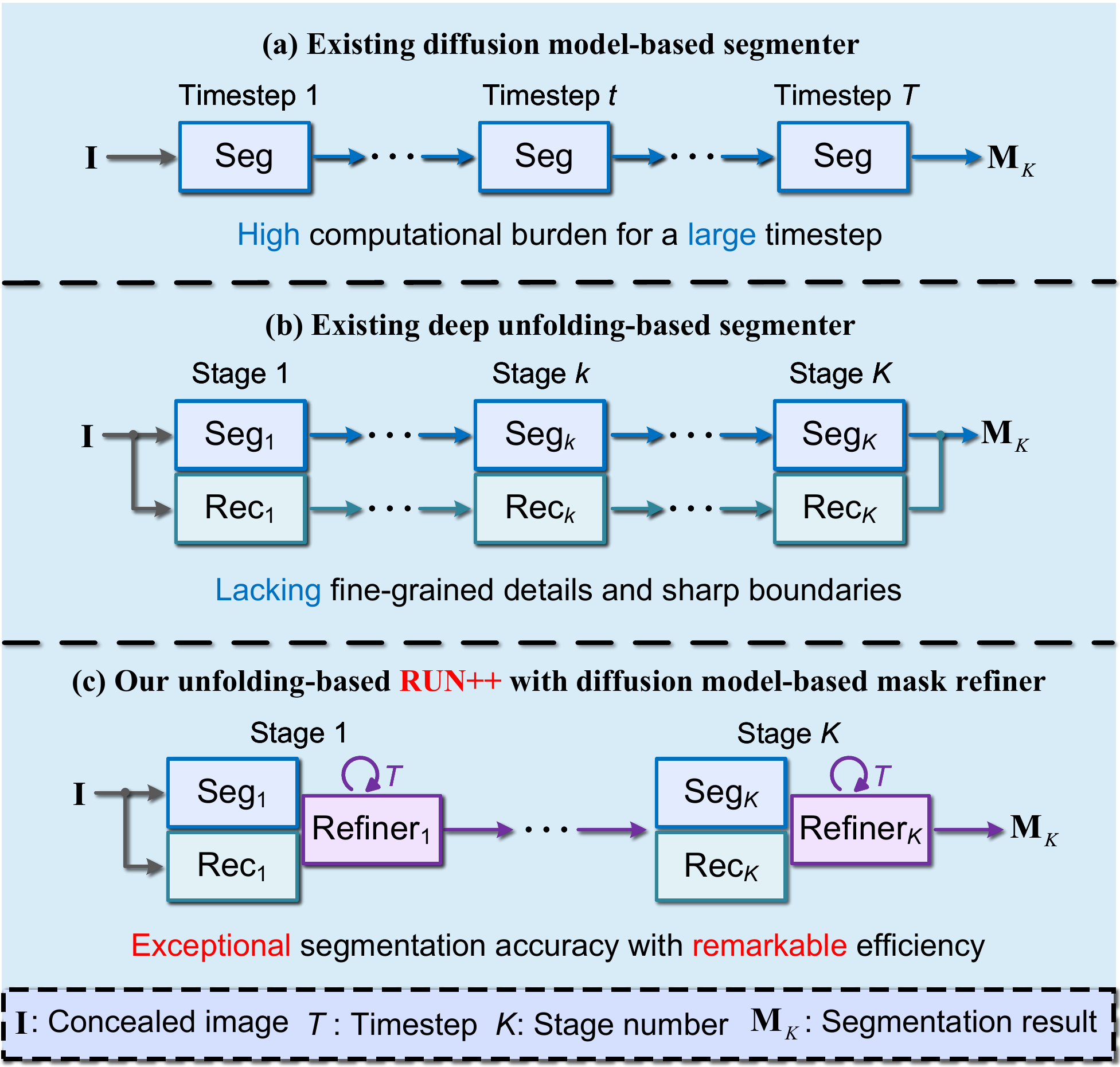}
	\captionof{figure}{Architecture our RUN++. (a)-(c) are the structural comparisons between existing diffusion model-based and deep unfolding-based segmenters~\cite{zhao2025focusdiffuser,sun2025conditional,he2025run} and our unfolding-based RUN++ with diffusion model-based mask refiner. 
    }
	\label{fig:Intro}
	\vspace{-5mm}
\end{figure}
\IEEEPARstart{C}{oncealed} Visual Perception (CVP) represents a fundamental computer vision challenge wherein objects are intrinsically blended with their surroundings, severely limiting the availability of discriminative cues. This problem is pervasive across a range of critical applications, including camouflaged object detection~\cite{fan2020camouflaged}, medical image analysis such as polyp~\cite{he2023weaklysupervised} and tubular object segmentation~\cite{he2025run}, and the detection of transparent objects~\cite{he2023strategic}.

The inherent ambiguity of CVP tasks renders them largely intractable for traditional methods that rely on hand-crafted features and manually designed segmentation rules~\cite{he2019image}. While early deep learning models advanced the field by leveraging powerful feature extractors, their common focus on foreground identification—exemplified by methods like SINet~\cite{fan2020camouflaged}—led them to neglect the rich contextual information available in the background. As illustrated in~\cref{fig:Intro,fig:Radar}, this oversight is a primary cause of suboptimal performance, as it leaves the model without the necessary cues to resolve ambiguity.

\begin{figure}[!h]
\setlength{\abovecaptionskip}{0cm}
	\centering
	\includegraphics[width=\linewidth]{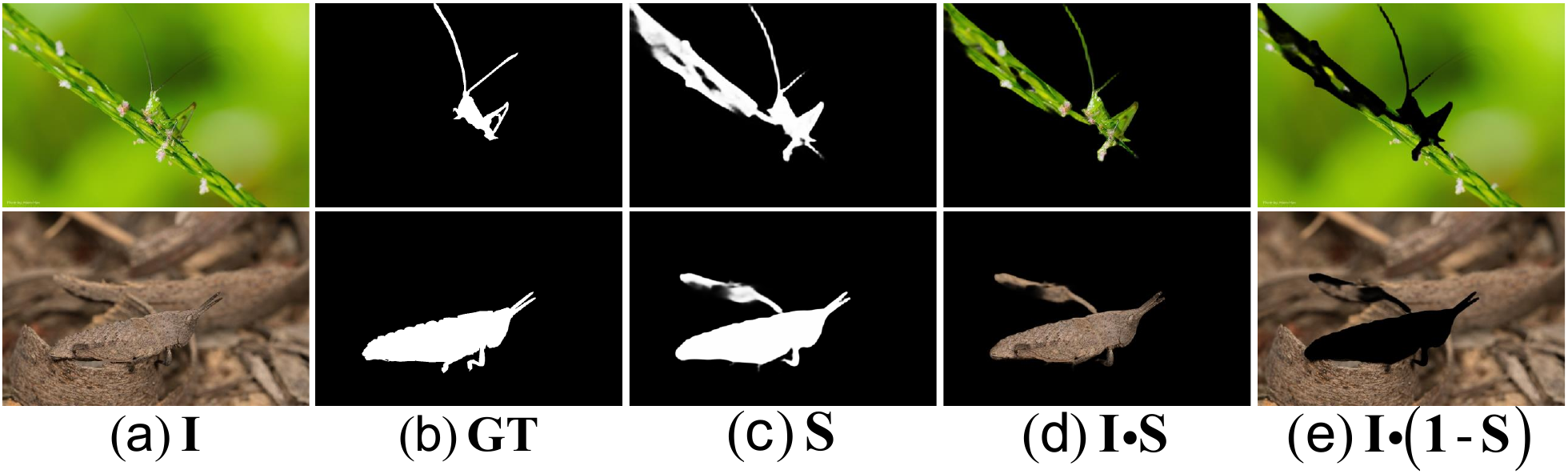}
	\caption{Correspondence between uncertainties in the mask domain and distortions in the RGB domain.
    ${\mathbf{I}}$ is the concealed image and ${\mathbf{S}}$ is the segmentation mask.
    }
	\label{fig:Recon}
	\vspace{-4mm}
\end{figure}

More recent algorithms, such as FEDER~\cite{He2023Camouflaged}, improve by employing reversible strategies—simultaneously modeling the foreground and background—at the mask level. 
This enhances the network's ability to identify subtle discriminative cues by focusing on regions of uncertainty. However, a key limitation persists: the potential of reversible modeling in the RGB domain remains unexplored. 
This is a missed opportunity, as uncertainties in a perception task often manifest as tangible color and texture distortions in the underlying RGB image (as depicted in~\cref{fig:Recon} (d) and (e)). 
Rectifying these RGB-level distortions can, in turn, help to resolve uncertainty in the mask.

This observation inspires us to turn to Deep Unfolding Networks (DUNs) as a principled way to apply reversible strategies across both mask and RGB domains. We propose integrating this approach with generative refinement in a novel architecture, termed RUN++, which is designed to resolve the distinct predictions arising from these two domains. The foundation of RUN++ is a mathematical optimization problem that we formulate for the CVP task, incorporating a novel foreground-background separation model to reduce segmentation uncertainty. We then unfold the iterative optimization solution for this model into a multistage deep network, making the original fixed hyperparameters learnable.

Each stage of RUN++ features three purpose-driven modules: a Concealed Object Region Extraction (CORE) module applies reversible modeling in the mask domain to identify primary object regions; a Context-Aware Region Enhancement (CARE) module extends this principle to the RGB domain to foster a more accurate foreground-background separation; and a Finetuning Iteration via Noise-based Enhancement (FINE) module introduces a stage-wise Bernoulli diffusion model (BDM) as a final refinement step. Uniquely, this BDM is applied in a targeted manner, refining only the uncertain regions of the segmentation mask identified by the preceding modules. As the network stages progress, RUN++ incrementally refines its foreground-background model in both the mask and RGB domains, directing its attention to ambiguous regions to progressively reduce false positives and negatives.

This design creates a powerful symbiosis: the unfolding network provides a high-quality uncertainty prior for the diffusion model, drastically reducing its computational cost, while the diffusion model restores fine-grained details, allowing the framework to achieve exceptional accuracy with efficiency.

Furthermore, we demonstrate the broad applicability of our RUN++'s core principle. We first extend the framework to create a degradation-resistant CVP system, a practical adaptation that maintains high performance in challenging real-world conditions like low-light or low-resolution. More broadly, we generalize this into a unified bi-level collaborative optimization (BLCO) framework. This framework moves beyond simple sequential pipelines by formalizing the synergistic, co-dependent relationship between low-level restoration and high-level perception, enhancing the robustness of high-level applications in degradation conditions and providing a principled blueprint for a new class of collaborative vision systems.

Our contributions are summarized as follows:

(1) We propose RUN++, a novel framework that, to our knowledge, is the first to address the CVP problem by synergizing the structural advantages of a deep unfolding network with the generative power of a diffusion model.

(2) RUN++ proposes a novel CVP model designed for uncertainty elimination, featuring the CORE and CARE modules that integrate model-based optimization to enable reversible modeling of the foreground and background.

(3) We introduce the FINE module, a targeted refiner that employs a stage-wise Bernoulli diffusion model to resolve segmentation uncertainties, creating a novel and efficient synergy between the deep unfolding network and diffusion model.

(4) We introduce a degradation-resistant adaptation of RUN++ and generalize its principles into a unified BLCO vision strategy that combines low-level and high-level tasks for robust high-level applications in degradation conditions.

(5) Extensive experiments conducted across five distinct CVP settings (encompassing 14 tasks), as well as on salient object detection, demonstrate the state-of-the-art performance, strong efficiency, and broad adaptability of RUN++.

A preliminary version of this work was published in~\cite{he2025run}. This paper presents a substantial extension with the following key contributions: 
First, we \textbf{propose RUN++, an advanced framework that integrates a stage-wise diffusion model into the original RUN architecture for mask refinement}. This novel design, to our knowledge, is the first to successfully integrate these two paradigms for concealed visual perception, incorporating the generative capacity of diffusion models while alleviating their computational burden.
Second, we \textbf{develop a methodology for degradation-resistant CVP and formulate a new bi-level optimization framework} to synergistically combine low-level and high-level vision algorithms, facilitating robust performance in challenging scenarios with visual degradation.
Furthermore, we \textbf{conduct a far more extensive experimental validation} across a wider range of CVP tasks, including weakly-supervised, semi-supervised, multimodal, and video segmentation. This ensures that our comparisons against \textbf{recently published algorithms} are comprehensive and rigorous.
We also \textbf{provide additional in-depth analyses and visualizations}, encompassing comprehensive ablation studies to offer deeper insights into the mechanisms of RUN++ and demonstrate its generalization capabilities.
Finally, we \textbf{present a thorough analysis of the intrinsic advantages of deep unfolding networks} for concealed visual perception, offering principles to guide the design of future DUN-based methods in this domain.

\section{Related Works}
\noindent \textbf{Concealed visual perception}. Deep learning has significantly advanced the field of concealed visual perception~\cite{xiao2024survey}, a domain encompassing several critical tasks such as camouflaged object detection~\cite{He2023Camouflaged}, polyp image segmentation~\cite{he2023weaklysupervised}, medical tubular object segmentation~\cite{he2025run}, and transparent object detection~\cite{xiao2023concealed}. Within this area, methods that employ reversible techniques to simultaneously model both foreground and background regions have emerged as a particularly promising research direction. For instance, PraNet~\cite{fan2020pranet} introduced a parallel architecture with a reverse attention mechanism to improve polyp segmentation accuracy.  
More recently, FEDER~\cite{He2023Camouflaged} leveraged both foreground and background masks, assisted by edge information, to jointly identify concealed objects, while BiRefNet~\cite{zheng2024bilateral} proposed a reconstruction module to refine segmentation masks using gradient information, enabling high-resolution output.
Despite these advances, existing methods have only applied reversible strategies at the mask level, leaving the potential of reversible modeling in the RGB domain largely unexplored. To address this limitation, we introduce RUN++, the first deep unfolding network designed for concealed visual perception. At each stage of the unfolding process, RUN++ employs two key modules, CORE and CARE, to reversibly model the segmentation process from both foreground and background perspectives and uses a FINE module to restore those high-uncertainty regions. This significantly improves segmentation accuracy by holistically modeling the object and its context and highlighting those uncertainty regions for diffusion refinement.

\noindent \textbf{Diffusion model-based object segmentation}. Methods for diffusion-based object segmentation can be broadly categorized into two types: generative segmentation~\cite{zhao2025focusdiffuser,sun2025conditional} and iterative refinement~\cite{wang2023segrefiner}.
Generative segmentation methods reformulate the task as a conditional generation process, where the final segmentation mask is progressively denoised from random noise, conditioned on the input image. In the CVP task, works such as FocusDiff~\cite{zhao2025focusdiffuser} and CamoDiff~\cite{sun2025conditional} directly generate segmentation results by designing specific networks and conditioning mechanisms. Although performing well in certain scenarios, they typically require generating the mask from scratch, which incurs significant computational overhead and makes it difficult to incorporate fine-grained priors.
Iterative refinement methods offer a more efficient and flexible approach by positioning the diffusion model as a segmentation refiner. These methods first utilize an existing segmentation network to produce an initial, potentially flawed, coarse mask. Subsequently, this coarse mask is used as the starting point for a diffusion process, where a few reverse denoising steps are applied to correct errors, sharpen boundaries, and complete details. SegRefiner~\cite{wang2023segrefiner} has validated the effectiveness of this coarse-to-fine strategy, demonstrating the immense potential of diffusion models for improving segmentation quality.
However, such refinement strategies remain underexplored in the highly challenging CVP domain, given the inherent object-background similarity. To bridge this gap, our work leverages the complementary strengths of deep unfolding networks and diffusion models. We introduce the FINE module, a novel and efficient refiner. Unlike general-purpose approaches, FINE employs a Bernoulli diffusion model to operate only on highly uncertain regions. These regions are intelligently identified from the distinct outputs of our CORE and CARE modules, providing a high-quality uncertainty prior for the diffusion process. This targeted approach allows us to harness the powerful generative capabilities of the diffusion model to recover fine-grained details in ambiguous areas, all while avoiding the prohibitive computational cost of full-image refinement.

\noindent \textbf{Deep unfolding network}. DUNs have emerged as a powerful technique, primarily in low-level vision, by integrating traditional model-based optimization algorithms with the feature-learning power of deep networks~\cite{he2023degradation,fang2024real}. This hybrid approach offers enhanced interpretability compared to purely data-driven methods. For instance, DeRUN~\cite{he2023degradation} introduced a degradation-resistant model for image fusion that was unfolded into a network to balance performance and efficiency, while CoRUN~\cite{fang2024real} proposed a DUN for dehazing based on the atmospheric scattering model, explicitly modeling the physical properties of haze.
However, the application of DUNs in high-level vision has remained limited. This is primarily due to the inherent difficulty of formulating explicit mathematical models for high-level vision tasks.
In this work, we bridge this gap by first proposing a novel optimization model for concealed visual perception designed to reduce segmentation uncertainty. 
We then introduce RUN++, a deep unfolding network derived from this model. 
By synergizing a classic optimization solution with deep networks, RUN++ achieves superior segmentation accuracy, demonstrating the significant potential of DUNs for advancing concealed visual perception. Furthermore, this work pioneers a new methodology for CVP by being the first to synergize the structural advantages of a deep unfolding network with the generative power of a diffusion model. This unique synergy enables highly accurate mask refinement while circumventing the prohibitive computational costs typically associated with diffusion processes.

\begin{figure*}[h]
\setlength{\abovecaptionskip}{0cm}
	\centering
	\includegraphics[width=\linewidth]{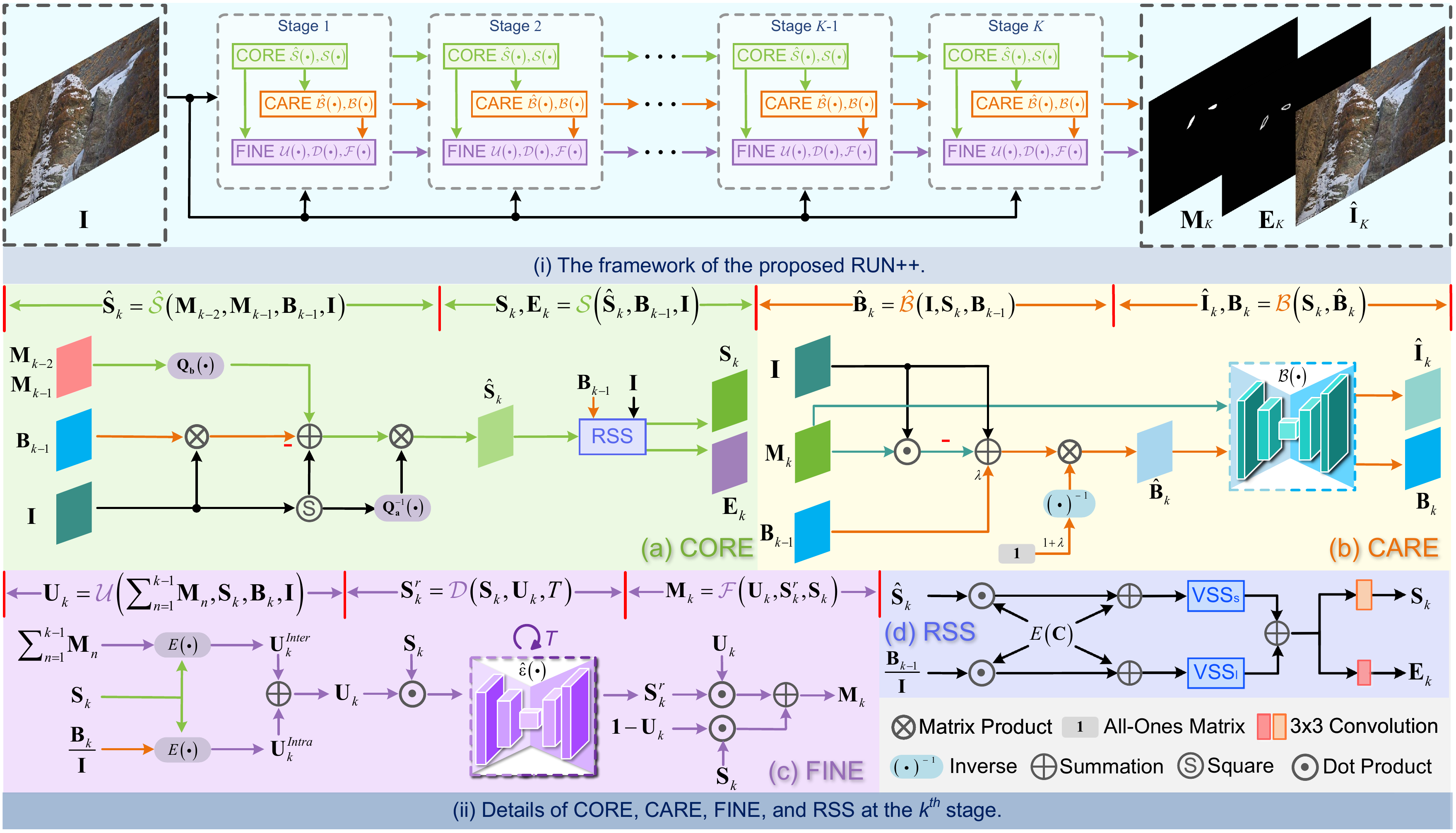}
	\caption{Framework of RUN++. In (a) and (b) of panel (ii), the connections within $\hat{\mathcal{S}}(\cdot)$ and $\hat{\mathcal{B}}(\cdot)$ are derived strictly based on mathematical principles, thus enhancing interpretability. In (c), $E(\bigcdot)$ means entropy and $T$ indicates the number of iterative diffusion time steps. For clarity, we replace certain redundant details with simple annotations, such as $\mathbf{Q}_\mathbf{a}$ and $\mathbf{Q}_\mathbf{b}$.
    }
	\label{fig:Framework}
	\vspace{-5mm}
\end{figure*}

\section{Methodology}
This section details the systematic development of our methodology, progressing from a foundational optimization model to a complete deep network framework.

First, in~\cref{Sec:COSM}, we formulate a novel optimization model for concealed visual perception (CVP). Subsequently, in~\cref{Sec:Optimization}, we derive an efficient optimization strategy for this model based on the proximal gradient algorithm. The iterative steps of this algorithm are then unfolded in~\cref{Sec:DUN} to construct a multi-stage deep network, termed RUN++. Finally, in~\cref{sec:BLO}, we demonstrate the versatility of RUN++ by adapting it for degradation-resistant CVP and generalizing its core principles into a unified bi-level optimization system.

\subsection{CVP Model} \label{Sec:COSM}
A concealed image $\mathbf{I}$ can be conceptually decomposed into its constituent foreground $\mathbf{F}$ and background $\mathbf{B}$ components:
\begin{equation}\label{Eq:Separation}
\mathbf{I}=\mathbf{F}+\mathbf{B}.
\end{equation}

Based on this principle, separating the foreground and background can be framed as an optimization problem. A foundational objective function is formulated as:
\begin{equation}\label{Eq:BasicModel}
   L\left(\mathbf{F}, \mathbf{B}\right)=\frac{1}{2}\|\mathbf{I}-\mathbf{F}-\mathbf{B}\|_2^2 + \beta\psi(\mathbf{F}) + \lambda \phi(\mathbf{B}),
\end{equation}
where the first term ensures data fidelity and utilize a $\ell_2$-norm $\|\bigcdot\|_2$ for smooth. $\psi(\mathbf{F})$ and $\phi(\mathbf{B})$ are regularization terms for the foreground $\mathbf{F}$ and background $\mathbf{B}$, respectively, with $\beta$ and $\lambda$ as their corresponding trade-off parameters. 

For segmentation tasks, our primary interest lies in the segmentation mask $\mathbf{S}$, where the foreground can be expressed as $\mathbf{F}= \mathbf{I}\bigcdot\mathbf{S}$ ($\bigcdot$ denotes element-wise multiplication). Substituting this into \cref{Eq:BasicModel}, the objective function becomes:
\begin{equation}\label{Eq:BasicModel1}
   L\left(\mathbf{S}, \mathbf{B}\right)=\frac{1}{2}\|\mathbf{I}- \mathbf{I}\bigcdot\mathbf{S} -\mathbf{B}\|_2^2 + \mu\varphi(\mathbf{S}) + \lambda \phi (\mathbf{B}),
\end{equation}
where $\varphi(\mathbf{S})$ and $\mu$ are the regularization term and trade-off parameter for the mask $\mathbf{S}$. However, due to the intrinsic ambiguity of concealed objects and the diversity of their surroundings, hand-crafting effective regularization terms for $\mathbf{S}$ and $\mathbf{B}$ is challenging. Hence, we propose to learn these priors implicitly using deep networks in a data-driven manner.

To further refine the segmentation and reduce pixel-level uncertainty, we introduce an additional residual sparsity constraint $\mathcal{T}(\bigcdot)$. This yields our final objective function:
\begin{equation}\label{Eq:FinalModel}
\begin{aligned}
       L\left(\mathbf{S}, \mathbf{B}\right)&=\frac{1}{2}\|\mathbf{I}- \mathbf{I}\bigcdot\mathbf{S} -\mathbf{B}\|_2^2 + \mu\varphi(\mathbf{S})\\
       & + \lambda \phi (\mathbf{B})
       +\alpha\mathcal{T}\left(\mathbf{w}\bigcdot\left(\mathbf{S}-\widetilde{\mathbf{S}}\right)\right),
\end{aligned}
\end{equation}
where $\mathcal{T}(\bigcdot)$ is an $\ell_1$-norm promoting sparsity, weighted by $\alpha$.
The term $\widetilde{\mathbf{S}}$ represents a refined mask after an uncertainty-removal mapping, and $\mathbf{w}$ is the attention map that focuses the constraint on high-confidence regions. For a pixel located at $(i,j)$, $\widetilde{\mathbf{S}}_{(i,j)}$ and $\mathbf{w}_{(i,j)}$ can be defined as follows:
\begin{equation}
\begin{aligned}
    \widetilde{\mathbf{S}}_{(i,j)}=&
\left\{
\begin{array}{ll}
    0.1  &\mathbf{S}_{(i,j)} \in [0.1,0.4),\\
    0.9  &\mathbf{S}_{(i,j)} \in (0.6,0.9],\\
    \mathbf{S}_{(i,j)} & \text{Otherwise}.
\end{array}\right. \\
\mathbf{w}_{(i,j)}=&
\left\{
\begin{array}{ll}
    0 \ \ \ \ \ \  &\mathbf{S}_{(i,j)} \in [0.4,0.6],\\
    1  & \text{Otherwise}.
\end{array}\right.
\end{aligned}
\end{equation}

This design encourages the network to produce decisive segmentation masks. Following the practice of~\cite{he2023weaklysupervised}, pixels with highly ambiguous values in the range $[0.4, 0.6]$ are excluded from this constraint via the attention map $\mathbf{w}$. The target values in $\widetilde{\mathbf{S}}$ are set to 0.1 and 0.9 (instead of 0 and 1) to provide greater optimization flexibility.

\subsection{Model Optimization} \label{Sec:Optimization}\vspace{-1mm}
We solve the objective function in~\cref{Eq:FinalModel} using the proximal gradient algorithm~\cite{fang2024real}, which finds the optimal mask $\mathbf{S}^*$ and background $\mathbf{B}^*$ by alternating updating each variable. We present the iterative solution for the $k^{th}$ stage ($1\leq k\leq K$).

\subsubsection{Update Rule for $\mathbf{S}_k$} 
The subproblem for updating the foreground mask $\mathbf{S}_k$ is formulated as follows:
\begin{equation}\label{Eq:SolutionM1}
\begin{aligned}
       \mathbf{S}_k = \arg \underset{\mathbf{S}}{\min } \ &\frac{1}{2}\|\mathbf{I}- \mathbf{I}\bigcdot\mathbf{S} -\mathbf{B}_{k-1}\|_2^2 + \mu\varphi(\mathbf{S})\\
       & +\alpha\mathcal{T}\left(\mathbf{w}_k\bigcdot\left(\mathbf{S}-\widetilde{\mathbf{S}}_k\right)\right).
\end{aligned}
\end{equation} 

The solution involves two main steps. First, a gradient descent step is performed to find an intermediate estimate $\hat{\mathbf{S}}_k$. Second, a proximal mapping step refines this estimate:
\begin{equation}\label{Eq:SolutionM2}
       \mathbf{S}_k = \text{prox}_\varphi\left(\mathbf{B}_{k-1},\hat{\mathbf{S}}_k\right),
\end{equation} 
where $\mathbf{B}_0$ and $\mathbf{S}_0$ are initialized as zero matrices. The variable $\hat{\mathbf{S}}_k$ is found by solving the following objective function:
\begin{equation}\hspace{-2mm}\label{Eq:SolutionM3}
\begin{aligned}
       \hat{\mathbf{S}}_k = \arg \underset{\hat{\mathbf{S}}}{\min } \ &\frac{1}{2}\|\mathbf{I}- \mathbf{I}\bigcdot \hat{\mathbf{S}} -\mathbf{B}_{k-1}\|_2^2 + \frac{\mu}{2}\|\hat{\mathbf{S}}-\mathbf{S}_{k-1}\|_2^2\\
       & +\alpha\mathcal{T}\left(\mathbf{w}_k\bigcdot\left(\hat{\mathbf{S}}-\widetilde{\mathbf{S}}_k\right)\right).
\end{aligned}
\end{equation} 

Note that $\mathbf{w}_k$ and $\widetilde{\mathbf{S}}_k$ are constructed based on ${\mathbf{S}_{k-1}}$. To create a differentiable objective, we approximate the non-differentiable $\ell_1$-norm term using a first-order Taylor expansion~\cite{goldstein1977optimization}. This allows us to derive a closed-form solution for $\hat{\mathbf{S}}_k$.
Following the practice of~\cite{goldstein1977optimization}, we approximate $\mathcal{T}(\mathbf{w}_k\bigcdot(\hat{\mathbf{S}}-\widetilde{\mathbf{S}}_k))$ at the $(k-1)^{th}$ iteration (for simplicity, we let $\mathbf{P}=\mathbf{w}_k\bigcdot(\hat{\mathbf{S}}-\widetilde{\mathbf{S}}_k)$), expressed as follows:
\begin{equation}\hspace{-2mm}\label{Eq:SolutionM4}
      \mathcal{T}\left(\mathbf{P}\right) \approx \dot{\mathcal{T}}\left(\mathbf{P,P_{k-1}}\right), 
\end{equation} 
where 
\begin{equation}\hspace{-2mm}\label{Eq:SolutionM5}
 \dot{\mathcal{T}}(\mathbf{P,P_{k-1}}\!) \!:=\! \frac{L_{\mathcal{T}}}{2}\|\mathbf{P}\!-\!\mathbf{P}_{k-1} \!+\! \frac{1}{L_{\mathcal{T}}}\nabla\mathcal{T}\!\left(\mathbf{P}_{k-1}\!\right)\!\|_2^2\!+\!C_\mathcal{T},
\end{equation}
where $L_{\mathcal{T}}$ is the Lipschitz constant. $\nabla\mathcal{T}(\mathbf{P}_{k-1})$ is the Lipschitz continuous gradient function of $\mathcal{T}(\mathbf{P}_{k-1})$ with $C_\mathcal{T}$, a positive constant that can be omitted in optimization. 
\setlength{\textfloatsep}{4pt}
\begin{algorithm}[t]
	\caption{Proposed RUN++ Framework.}
	\label{Alg:RUN}
	\textbf{Input}: concealed image $\mathbf{I}$, stage number $K$, time step $T$ \\
	\textbf{Output}: concealed object mask $\mathbf{M}_K$, concealed object edge $\mathbf{E}_K$, reconstructed concealed image $\hat{\mathbf{I}}_K$
	\begin{algorithmic}[1]
		\State Zero initialization for $\mathbf{S}_0$, $\mathbf{B}_0$
		\For{each stage $k\in \left[1,K\right]$} 
            \State $\hat{\mathbf{S}}_k = \hat{\mathcal{S}}\left(\mathbf{B}_{k-1}, \mathbf{M}_{k-1},  \mathbf{M}_{k-2}, \mathbf{I}\right)$,  \Comment{\cref{Eq:SOFSMhat}}
            \State $\mathbf{S}_k, \mathbf{E}_k  = \mathcal{S}\left( \mathbf{B}_{k-1}, \hat{\mathbf{S}}_k, \mathbf{I}\right)$, \Comment{\cref{eq:CORE2}}
		\State $\hat{\mathbf{B}}_k =\hat{\mathcal{B}}\left(\mathbf{B}_{k-1}, \mathbf{S}_k, \mathbf{I}\right),$  \Comment{\cref{Eq:SOFSBhat}}
        \State ${\mathbf{B}}_k, \hat{\mathbf{I}}_k = {\mathcal{B}}\left(\hat{\mathbf{B}}_{k}, \mathbf{I}\bigcdot\mathbf{S}_k\right),$ \Comment{\cref{eq:CARE2}}
        \State $\mathbf{U}_k=\mathcal{U}\left(\sum_{n=1}^{k-1} \mathbf{M}_n,\mathbf{S}_k,\mathbf{B}_k, \mathbf{I}\right),$ \Comment{\cref{Eq:Uncertainty}}
        \State $\mathbf{S}^r_k = \mathcal{D}\left(\mathbf{S}_k,\mathbf{U}_k,T\right),$   \Comment{\cref{eq:BDM}}  
        \State $\mathbf{M}_k = \mathcal{F}\left(\mathbf{U}_k, \mathbf{S}^r_k, \mathbf{S}_k\right). 
        $ \Comment{\cref{eq:FINE}} 
		\EndFor
	\end{algorithmic}
\end{algorithm}
Substituting into~\cref{Eq:SolutionM3}, we obtain the following equations:
\begin{equation}\hspace{-3mm}\label{Eq:SolutionM6}
\begin{aligned}
       \hat{\mathbf{S}}_k = &\frac{1}{2}\|\mathbf{I}- \mathbf{I}\bigcdot \hat{\mathbf{S}} -\mathbf{B}_{k-1}\|_2^2 + \frac{\mu}{2}\|\hat{\mathbf{S}}-\mathbf{S}_{k-1}\|_2^2\\
       & +\frac{\alpha L_{\mathcal{T}}}{2}\|\mathbf{P}\!-\!\mathbf{P}_{k-1} \!+\! \frac{1}{L_{\mathcal{T}}}\nabla\mathcal{T}\left(\mathbf{P}_{k-1}\right)\|_2^2.
\end{aligned}
\end{equation}

\cref{Eq:SolutionM6} can be solved directly by equating its derivative to zero. The closed-form solution is 
\begin{equation}\hspace{-3mm}\label{Eq:SolutionMFinal}
       \hat{\mathbf{S}}_k \!= \!\left(\mathbf{Q_a}\right)^{-1} \!\left(\mathbf{Q_b}{\mathbf{S}}_{k-1} \!+\! \mathbf{I}^2 \!-\!\mathbf{C B_{k-1}} \!+\! \mathbf{Q_c}\right)\!,
\end{equation} 
where $\mathbf{Q_a}$, $\mathbf{Q_b}$, $\mathbf{Q_c}$, and $\mathbf{Q_d}$ are intermediate matrices composed of terms from the optimization problem. $\mathbf{Q_a}=\mathbf{I}^2+L_{\mathcal{T}}\mathbf{w}_k^2 +\mu \mathbf{1}$, $\mathbf{1}$ is an all-ones matrix, $\mathbf{Q_b}=\alpha L_{\mathcal{T}}\mathbf{w}_k \mathbf{w}_{k-1}+\mu \mathbf{1}$, 
$\mathbf{Q_c}=\alpha L_{\mathcal{T}}\mathbf{w}_k (\mathbf{w}_k\bigcdot \widetilde{\mathbf{S}}_k - \mathbf{Q_d}) - \alpha \mathbf{w}_k \nabla \mathcal{T}(\mathbf{w}_{k-1}\bigcdot {\mathbf{S}}_{k-1}-\mathbf{Q_d})$, and $\mathbf{Q_d}=\mathbf{w}_{k-1}\bigcdot \widetilde{\mathbf{S}}_{k-1}$. 

\subsubsection{Update Rule for $\mathbf{B}_k$} 
Similarly, the background $\mathbf{B}_k$ is updated by solving the following objective function:
\begin{equation}\label{Eq:SolutionB1}
       \mathbf{B}_k=\arg \underset{\mathbf{B}}{\min } \frac{1}{2}\|\mathbf{I}- \mathbf{I}\bigcdot\mathbf{S}_k -\mathbf{B}\|_2^2 + \lambda \phi (\mathbf{B}).
\end{equation}

This also involves a gradient step to find $\mathbf{B}_k$, followed by a proximal step:
\begin{equation}\hspace{-2mm}\label{Eq:SolutionB2}
\setlength{\abovedisplayskip}{0pt}
\setlength{\belowdisplayskip}{0pt}
       \hat{\mathbf{B}}_k=\arg \underset{\hat{\mathbf{B}}}{\min } \frac{1}{2}\|\mathbf{I}- \mathbf{I}\bigcdot\mathbf{S}_k - \hat{\mathbf{B}}\|_2^2 + \frac{\lambda}{2} \|\hat{\mathbf{B}} - \mathbf{B}_{k-1}\|^2_2,
\end{equation} 
\begin{equation}\label{Eq:SolutionB3}
\setlength{\abovedisplayskip}{0pt}
\setlength{\belowdisplayskip}{-5pt}
       \mathbf{B}_k = \text{prox}_\phi (\hat{\mathbf{B}}_{k},\mathbf{S}_k ).
\end{equation} 

The closed-form solution for the intermediate background estimate $\hat{\mathbf{B}}_k$ can be calculated similarly:
\begin{equation}\hspace{-2mm}\label{Eq:SolutionBFinal}
       \hat{\mathbf{B}}_k= \left(\left(1+\lambda \right)\mathbf{1}\right)^{-1}\left(\lambda\mathbf{B}_{k-1}+\mathbf{I}- \mathbf{I}\bigcdot\mathbf{S}_k \right).
\end{equation}

\subsection{RUN++} \label{Sec:DUN}
We unfold the iterative optimization process into a multi-stage network, named RUN++. Each stage mirrors one iteration of the optimization-based algorithm. As shown in~\cref{fig:Framework}, a stage consists of three modules: the Concealed Object Region Extraction (CORE), Context-Aware Region Enhancement (CARE), and Finetuning Iteration via Noise-based Enhancement (FINE) modules. CORE aims to segment those \textbf{core} concealed objects, CARE \textbf{cares} on fostering better foreground-background separation, while FINE introduces a Bernoulli diffusion model to refine the segmentation based on the uncertainty regions highlighted by reversible outputs. This helps generate more accurate masks, denoted as $\mathbf{M}$, with fewer uncertain regions and \textbf{finer} details. The detailed variable update process for each stage is shown in~\cref{Alg:RUN}.

\subsubsection{Concealed Object Region Extraction (CORE) Module}
CORE corresponds to the update rule for the segmentation result ${\mathbf{S}}_k$ in~\cref{Eq:SolutionM2,Eq:SolutionMFinal}. It first computes the intermediate mask $\hat{\mathbf{S}}_k$ using a network layer $\hat{\mathcal{S}}(\bigcdot)$, which implements the closed-form solution from~\cref{Eq:SolutionMFinal} but with learnable parameters to enhance adaptability, formulated as:
\begin{equation}\hspace{-3mm}\label{Eq:SOFSMhat}
\begin{aligned}
       \hat{\mathbf{S}}_k &=\! \hat{\mathcal{S}}\left(\mathbf{B}_{k-1}, \mathbf{M}_{k-1}, \mathbf{M}_{k-2}, \mathbf{I}\right)\!, \\ &=\!\left(\mathbf{Q_a}\right)^{-1} \!\left(\mathbf{Q_b}{\mathbf{M}}_{k-1} \!+\! \mathbf{I}^2 \!-\!\mathbf{CB_{k-1}} \!+\! \mathbf{Q_c}\right)\!,
\end{aligned}
\end{equation} 
where $\mathbf{M}_{k-1}$ and $\mathbf{M}_{k-2}$ are corresponding to the masks refined by the FINE module at the $\{k-1\}^{th}$ and $\{k-2\}^{th}$ stages.
This initial mask $\hat{\mathbf{S}}_k$ is then refined by the Reversible State Space (RSS) module, $RSS(\bigcdot)$, which leverages non-local information for accurate feature extraction. The RSS module contains two Visual State Space (VSS) $VSS(\bigcdot)$ blocks~\cite{Liu2024vmamba} with different receptive fields: a small-field VSS locally refines object boundaries from a foreground perspective, while a large-field VSS globally identifies missed regions from a background perspective. This dual-perception mechanism produces the refined mask $\mathbf{S}_k$ and an auxiliary edge output $\mathbf{E}_k$, which are proven to facilitate the generation of more accurate segmentation results~\cite{He2023Camouflaged}. They are defined as follows:
\begin{equation}\hspace{-3mm}\label{eq:CORE2}
\begin{aligned}
    \mathbf{S}_k,\mathbf{E}_k &\!=\!  \mathcal{S}( \mathbf{B}_{k-1}\!, \hat{\mathbf{S}}_k,\mathbf{I}) \!= \! RSS ( \mathbf{B}_{k-1},\hat{\mathbf{S}}_k,\mathbf{I}), \\
    &\!=\! conv3(VSS_s(E(\mathbf{I})\bigcdot\hat{\mathbf{S}}_k + E(\mathbf{I})) \\
    &\!+\! VSS_l (E(\mathbf{I})\bigcdot(\mathbf{B}_{k-1}/\mathbf{I}) + E(\mathbf{I}) )),
\end{aligned}
\end{equation}
where $conv3$ represents a $3\times3$ convolution. $VSS_s(\bigcdot)$ and $VSS_l(\bigcdot)$ correspond to the VSS blocks with small and large kernel sizes, thus having varying perception fields. We omit the detailed description of VSS for brevity.

Recognizing that complex segmentation tasks like CVP are heavily reliant on high-level semantic information that is difficult to capture from raw pixels, the RSS module operates on deep feature representations. We therefore extract rich semantic features $E(\mathbf{I})$ from the input image using a shared encoder (\textit{e.g.}, ResNet50~\cite{he2016deep}). By having the RSS module process these deep features instead of the raw image, our network can discern subtle yet critical discriminative patterns, leading to a more accurate and robust segmentation.

\subsubsection{Context-Aware Region Enhancement (CARE) Module} 
The CARE module updates the background estimation, mirroring the update for ${\mathcal{B}}(\bigcdot)$. It first calculates an intermediate background $\hat{\mathbf{B}}_k$ using a learnable network layer $\hat{\mathcal{B}}(\bigcdot)$ based on~\cref{Eq:SolutionBFinal}, expressed as: 
\begin{equation}\hspace{-2mm}\label{Eq:SOFSBhat}
\begin{aligned}
    \hat{\mathbf{B}}_k&=\hat{\mathcal{B}}\left(\mathbf{B}_{k-1}, \mathbf{S}_k, \mathbf{I}\right), \\
    &= \left(\left(1+\lambda \right)\mathbf{1}\right)^{-1}\left(\lambda\mathbf{B}_{k-1}+\mathbf{I}- \mathbf{I}\bigcdot\mathbf{S}_k \right).
\end{aligned}
\end{equation} 

This layer is essentially a dynamic fusion of the previously estimated background and the reversed foreground.

This estimation is then refined by a U-shaped network $\mathcal{B}(\bigcdot)$~\cite{he2023HQG} to produce the final background $\hat{\mathbf{B}}_k$. However, since the CORE and CARE modules estimate the foreground and background separately, their interpretations of the concealed content may differ (see~\cref{fig:Recon}). This can lead to conflicting estimations, creating distortion-prone areas in a naively reconstructed image. To address this, the refinement network $\mathcal{B}(\bigcdot)$ is also tasked with generating a reconstructed image $\hat{\mathbf{I}}_k$, formulated as:
\begin{equation}\hspace{-2mm}
\begin{aligned}
    {\mathbf{B}}_k, \hat{\mathbf{I}}_k = {\mathcal{B}}\left(\hat{\mathbf{B}}_{k}, \mathbf{I}\bigcdot\mathbf{S}_k\right).
\end{aligned}\label{eq:CARE2}
\end{equation} 

By penalizing differences between $\hat{\mathbf{I}}_k$ and the original input $\mathbf{I}$, we enforce consistent judgments between the two modules. This alignment fosters better foreground-background separation and ultimately improves segmentation accuracy.

\subsubsection{Finetuning Iteration via Noise-based Enhancement (FINE) Module} 
The FINE module introduces a Bernoulli diffusion model (BDM)~\cite{sohl2015deep}, a discrete diffusion process well-suited for binary segmentation masks, to further refine the output segmentation result $\mathbf{S}_k$. 
This design establishes a novel synergy between two powerful iterative frameworks: DUN and the diffusion model. 
Although both are iterative, their underlying mechanisms are highly complementary. The DUN provides a high-quality initial mask for the diffusion process, which significantly reduces the number of reverse steps required for convergence and thereby mitigates the substantial computational overhead typically associated with diffusion models.
Conversely, the diffusion model’s exceptional capability for generating fine-grained details relieves the DUN from needing an excessive number of stages to achieve high precision. This reciprocal relationship ensures that the overall framework is both effective and efficient.

To enhance efficiency, rather than applying the BDM to the entire mask, we employ a targeted refinement strategy. An uncertainty map is generated to identify ambiguous regions, and the BDM is applied exclusively to these areas. This process is formulated as follows:
\begin{equation}\label{eq:FINE}
    \mathbf{M}_k = \mathcal{F}(\mathbf{U}_k, \mathbf{S}^r_k, \mathbf{S}_k) = \mathbf{U}_k \bigcdot \mathbf{S}^r_k + (\mathbf{1} - \mathbf{U}_k) \bigcdot \mathbf{S}_k,
\end{equation}
where $\mathbf{M}_k$ is the final mask for stage $k$. The refinement is guided by an uncertainty map $\mathbf{U}_k$, which is derived from the current mask $\mathbf{S}_k$. $\mathbf{S}^r_k$ represents the refined mask within the uncertain regions, processed by BDM $\mathcal{D}(\bigcdot)$, defined as:
\begin{equation}\label{eq:BDM}
    \mathbf{S}^r_k = \mathcal{D}(\mathbf{S}_k,\mathbf{U}_k,T),
\end{equation}
where $T$ is the number of diffusion time steps.

\noindent\textbf{Uncertainty map generation}. 
Our final uncertainty map, $\mathbf{U}_k$, is generated by combining two distinct forms of uncertainty: inter-stage and intra-stage. To compute these maps, we follow the common practice~\cite{he2023weaklysupervised,he2025segment} of using Shannon entropy, a method that is significantly more computationally efficient than network-based estimation approaches~\cite{yang2021uncertainty}.

For \textit{inter-stage uncertainty}, we aim to identify regions with ambiguous decisions across multiple stages of the network. This is achieved by fusing the segmentation outputs from all preceding $k$ stages. This approach ensures high recall (\textit{i.e.}, complete segmentation) by preserving areas that may have been inconsistently classified. The inter-stage uncertainty map, $\mathbf{U}^{Inter}_k$, is formulated by applying a pixel-wise entropy function, $E(\bigcdot)$, to the averaged masks:
\begin{equation}\hspace{-3mm}
\mathbf{U}^{Inter}_k = E(\mathbf{S}_k^{Inter}), \ \text{where} \ \mathbf{S}_k^{Inter} = \frac{1}{k}(\sum_{n=1}^{k-1} \mathbf{M}_n+\mathbf{S}_k).
\end{equation}

For \textit{intra-stage uncertainty}, the goal is to highlight regions where different modules within the same stage produce conflicting judgments. We compute this by combining the primary segmentation mask, $\mathbf{S}_k$, with the inverted output of the background-focused branch, $\mathbf{B}_k$. This focuses refinement on contested pixels, thereby ensuring high precision (\textit{i.e.}, accurate segmentation). The intra-stage map, $\mathbf{U}^{Intra}_k$, is calculated as:
\begin{equation}\hspace{-3mm}
    \mathbf{U}^{Intra}_k = E(\mathbf{S}_k^{Intra}), \ \text{where} \ \mathbf{S}_k^{Intra}=\frac{1}{2} ( \mathbf{S}_k + \mathbf{1} - \frac{\mathbf{B}_k}{\mathbf{I}} ).
\end{equation}

In the equations above, the function $E(p)=-p\log_2(p)-(1-p)\log_2(1-p)$ represents the pixel-wise binary entropy.

Finally, the two uncertainty maps are averaged to produce the comprehensive uncertainty map $\mathbf{U}_k$, which guides the refinement process. This entire operation is abstracted as $\mathcal{U}(\bigcdot)$:
\begin{equation}\label{Eq:Uncertainty}
    \mathbf{U}_k=\mathcal{U}(\sum_{n=1}^{k-1} \mathbf{M}_n,\mathbf{S}_k,\mathbf{B}_k, \mathbf{I})=\frac{1}{2}(\mathbf{U}^{Inter}_k+\mathbf{U}^{Intra}_k).
\end{equation}

\noindent\textbf{Bernoulli diffusion model}. 
The BDM is responsible for refining the segmentation mask within the uncertain regions. Our implementation employs a UNet-based architecture with ResNet50 serving as the denoiser. To enhance the perception of concealed objects, features extracted by the BDM's encoder are fused with the image features $E(\mathbf{I})$ from~\cref{eq:CORE2}.

During training, the goal of the \textit{forward process} is to construct a diffusion chain $q(\mathbf{m}_{1:T}|\mathbf{m}_0, \mathbf{S}^{GT}_k)$ that gradually corrupts an initial mask $\mathbf{m}_0$ towards a noisy state $\mathbf{m}_T$. This process is conditioned on the ground truth signal within the uncertain regions, defined as $\mathbf{S}^{GT}_k=\mathbf{U}_{GT}\bigcdot \mathbf{S}_k$. The binary map $\mathbf{U}_{GT}$ identifies pixels where the initial prediction $\mathbf{S}_k$ differs from the ground truth mask $\mathbf{GT}_s$. By iteratively adding Bernoulli noise over $T$ steps, each step in this forward process can be formulated as: 
\begin{equation}\label{eq:DM1}
    q(\mathbf{m}_{t}|\mathbf{m}_{t-1}, \mathbf{S}^{GT}_k) = \mathcal{B}_d((1-\beta_{t}) \mathbf{m}_{t-1}+\beta_{t}\mathbf{S}^{GT}_k).
\end{equation}
where $t=1, \cdots, T$, $\beta_t$ is a predefined noise schedule, and $\mathcal{B}_d$ is the Bernoulli distribution. Following~\cite{kingma2013auto}, this process can be expressed in a closed form:
\begin{equation}\hspace{-3mm}\label{eq:DM2}
q(\mathbf{m}_{t}|\mathbf{m}_{0}, \mathbf{U}_{GT}, \mathbf{S}_k) \!=\! \mathcal{B}_d(\mathbf{U}_{GT} \bigcdot (\bar{\alpha}_{t} \mathbf{U}_{GT} \!+\! (1-\bar{\alpha}_{t})\mathbf{S}_k)),
\end{equation}
where $\alpha_{t}=1-\beta_{t}$ and $\bar{\alpha}_{t}= {\prod_{i=1}^{t}} {\alpha}_{i}$. 
To implement uncertainty-aware noise injection, we reparameterize the noisy mask $\mathbf{m}_t$ using the exclusive OR operator $\oplus$:
\begin{equation}\label{eq:DM3}
 \mathbf{m}_t = \mathbf{m}_0 \oplus \bm{\epsilon}, \ \ \bm{\epsilon} \sim \mathcal{B}_d((1-\bar{\alpha}_{t}) | \mathbf{S}^{GT}_k - \mathbf{m}_0 |). 
\end{equation}

This uncertainty-aware modulation confines the diffusion process exclusively to the regions where the initial prediction was incorrect. This strategy leverages the BDM's generative capabilities while preventing it from corrupting already accurate regions—a critical consideration given the potential ambiguity between a concealed object and its background. The resulting Bernoulli posterior is formulated as:
\begin{equation}\label{eq:DM4}
q(\mathbf{m}_{t-1}|\mathbf{m}_{t},\mathbf{m}_{0}, \mathbf{S}^{GT}_k) = \mathcal{B}_d(\Phi_p(\mathbf{m}_{t},\mathbf{m}_{0},\mathbf{S}^{GT}_k)).
\end{equation}

Here, $\Phi_p$ is a function of its arguments as formulated in~\cite{Hoogeboom2021ArgmaxFA}, involving products of direct and inverted probabilities.

\setlength{\textfloatsep}{4pt}
\begin{algorithm}[t]
	\caption{RUN++ for Degradation-Resistant CVP.}
	\label{Alg:RUN++}
	\textbf{Input}: degraded concealed image $\mathbf{I}$, stage number $K$, time step $T$ \\
	\textbf{Output}: concealed object mask $\mathbf{M}_K$, concealed object edge $\mathbf{E}_K$, restored concealed image $\hat{\mathbf{I}}_K$
	\begin{algorithmic}[1]
		\State Zero initialization for $\mathbf{S}_0$, $\mathbf{B}_0$
            \State Initializing $\hat{\mathbf{I}}_0$ as $\mathbf{I}$
		\For{each stage $k\in \left[1,K\right]$} 
            \State $\hat{\mathbf{S}}_k = \hat{\mathcal{S}}\left(\mathbf{B}_{k-1}, \mathbf{M}_{k-1},  \mathbf{M}_{k-2}, \hat{\mathbf{I}}_{k-1}\right)$, 
            \State $\mathbf{S}_k, \mathbf{E}_k  = \mathcal{S}\left( \mathbf{B}_{k-1}, \hat{\mathbf{S}}_k, \hat{\mathbf{I}}_{k-1}\right)$, 
		\State $\hat{\mathbf{B}}_k =\hat{\mathcal{B}}\left(\mathbf{B}_{k-1}, \mathbf{S}_k, \hat{\mathbf{I}}_{k-1}\right),$ 
        \State ${\mathbf{B}}_k, \hat{\mathbf{I}}_k = \tilde{\mathcal{B}}\left(\hat{\mathbf{B}}_{k}, \hat{\mathbf{I}}_{k-1}\bigcdot\mathbf{S}_k, \hat{\mathbf{I}}_{k-1}\right),$ 
        \State $\mathbf{U}_k=\mathcal{U}\left(\sum_{n=1}^{k-1} \mathbf{M}_n,\mathbf{S}_k,\mathbf{B}_k, \hat{\mathbf{I}}_{k-1}\right),$ 
        \State $\mathbf{S}^r_k = \mathcal{D}(\mathbf{S}_k,\mathbf{U}_k,T),$ 
        \State $\mathbf{M}_k = \mathcal{F}(\mathbf{U}_k, \mathbf{S}^r_k, \mathbf{S}_k). 
        $ 
		\EndFor
	\end{algorithmic}
\end{algorithm}

During inference, the \textit{reverse process} iteratively refines the initial mask $\mathbf{S}_k$ within the uncertain regions identified by $\mathbf{U}_k$. Let this target sub-mask be $\mathbf{S}^U_k = \mathbf{U}_k \bigcdot \mathbf{S}_k$. Beginning with a random sample \( \mathbf{m}_T \sim \mathcal{B}_d(\mathbf{S}^U_k)\), the reverse process is a learned Markov chain that gradually denoises the mask from $\mathbf{m}_t$ back to $\mathbf{m}_{t-1}$:
\begin{equation}
    p_\theta(\mathbf{m}_{t-1}|\mathbf{m}_{t}, \mathbf{S}^U_k) = \mathcal{B}_d(\hat{\mu}(\mathbf{m}_t, t, \mathbf{S}^U_k)),
\end{equation}
where the refined mask mean $\hat{\mu}$ is estimated by the denoiser network $\hat{\epsilon}(\bigcdot)$:
\begin{equation}
    \hat{\mu}(\mathbf{m}_t, t, \mathbf{S}^U_k) = \Phi_p (\mathbf{m}_t, |\mathbf{m}_t - \hat{\epsilon}(\mathbf{m}_t,t,\mathbf{S}^U_k) |, \mathbf{S}^U_k). 
\end{equation}

After $T$ iterative denoising steps,
the process yields the final prediction $\hat{\mathbf{m}}_0$, which corresponds to the refined mask $\mathbf{S}^r_k$ in~\cref{eq:BDM}. Substituting $\mathbf{S}^r_k$ into \cref{eq:BDM} produces the final, enhanced segmentation mask $\mathbf{M}_k$ for the stage, which is expected to have higher accuracy and finer details.

As the stages progress, RUN++ incrementally facilitates reversible modeling of the foreground and background in both the mask and RGB domains, along with a refiner to further eliminate the foreground-background misalignment. This process systematically directs the network's attention to ambiguous regions, progressively reducing false-positive and false-negative errors. This iterative cycle culminates in a final segmentation that is robust, complete, and precise.

\subsubsection{Loss function} 
The total loss function is a composite objective that supervises the outputs of all three modules at each stage. 
It combines standard segmentation and edge losses~\cite{He2023Camouflaged}, a reconstruction fidelity term, and a term for training the diffusion model. The overall loss $L_t$ is defined as: 
\begin{equation}\hspace{-3mm}
\begin{aligned}
		L_t\!&=\!\sum_{k=1}^{K}\frac{1}{2^{K-k}}\!\left[L^w_{BCE}\!\left(\mathbf{M}_k,\mathbf{GT}_s\right)\!+\!L^w_{IoU}\!\left(\mathbf{M}_k,\mathbf{GT}_s\right)\right.\\
		& +\! L_{dice}\left(\mathbf{E}_k,\mathbf{GT}_e\right)+ \|\hat{\mathbf{I}}_k-\mathbf{I} \|^2_2+ L^{KL}_k \ ],
	\end{aligned}
\end{equation}
\begin{equation}\hspace{-3mm}
    L^{KL}_k = KL(q(\mathbf{m}_{t-1}|\mathbf{m}_t,\mathbf{m}_0,\mathbf{S}^U_k)||p_\theta(\mathbf{m}_{t-1}|\mathbf{m}_t, \mathbf{S}^U_k)),
\end{equation}
where $K$ is the stage number. 
$L_{BCE}^w$, $L_{IoU}^w$, and $L_{dice}$ are the weighted binary cross-entropy, weighted intersection-over-union, and dice losses. $\mathbf{GT}_s$ and $\mathbf{GT}_e$ are the ground truth of the segmentation mask and edge.

\subsection{Degradation-Resistant CVP and Bi-level Optimization}~\label{sec:BLO}
In this section, we extend the utility of our RUN++ framework in two significant directions. First, we adapt it for degradation-resistant concealed visual perception (CVP). Second, we generalize its core principles into a bi-level optimization framework for unified vision systems.

\noindent\textbf{Degradation-resistant CVP}. 
We now extend RUN++ to address complex scenarios involving specific degradations.  
To adapt RUN++ for such conditions, the generic reconstruction network $\mathcal{B}(\bigcdot)$ within the CARE module is replaced with a specialized restoration network $\tilde{\mathcal{B}}(\bigcdot)$ tailored to a specific degradation type. Consider two common and challenging scenarios: low-light and low-resolution imagery. Low-light conditions can exacerbate object concealment by reducing color contrast, while low-resolution inputs diminish discriminative cues by decreasing the number of effective pixels.

Our adaptation strategy is modular, employing a specialized model for each degradation type. 
For well-defined degradations, we integrate pre-trained networks, such as a dedicated illumination enhancement model like Reti-Diff~\cite{he2023reti} for low-light inputs or an image super-resolution model~\cite{hu2025iqpfr} for low-resolution ones. For images with unknown corruptions, a blind image restoration model~\cite{xia2023diffir} is utilized. 
This adaptation effectively transforms the generic reconstruction sub-task into a targeted restoration task. The goal is twofold: not only to resolve the conflicting estimations as in Eq.~\eqref{eq:CARE2}, but also to explicitly counteract the visual interference from the degradation, thereby ensuring robust segmentation.
During training, the restoration network is optimized via a dual-supervision scheme: an explicit constraint from the clean ground-truth image encourages high-fidelity restoration, while an implicit constraint from the primary segmentation loss guides the network to produce object-centric features. This dual objective ensures that the network not only generates visually pristine images but also enhances the most salient cues required for accurate downstream segmentation.

However, incorporating cutting-edge restoration algorithms into our multi-stage network can introduce a non-negligible computational burden, particularly as the number of stages increases. To mitigate this, we propose a coarse-to-fine restoration strategy across the network's stages. Lightweight restoration models are used in the initial stages to provide a preliminary enhancement at a low computational cost. In the final, decisive stages, more powerful restoration models are employed to maximize segmentation accuracy. 
This progressive approach is implemented by feeding the restored image from the previous stage, $\hat{\mathbf{I}}_{k-1}$, as an input to the current stage (with the initial input $\hat{\mathbf{I}}_0$ being the original degraded image $\mathbf{I}$). 
The newly restored image $\hat{\mathbf{I}}_k$ from CORE can be passed to the $(k+1)^{th}$ stage, providing progressively enhanced visual cues to directly guide the segmentation process. 
This requires a reformulation of the deep networks in the CARE module (line 7 in Algorithm~\ref{Alg:RUN++}) to accept $\hat{\mathbf{I}}_{k-1}$ as an additional input. 
It is important to note that this specific architecture is employed only for the degradation-resistant setting; it was intentionally avoided in our primary framework to prevent the network from learning a trivial solution, such as an identity mapping, on non-degraded images. The complete coarse-to-fine process is formally detailed in Algorithm~\ref{Alg:RUN++}.

\begin{figure}[t]
\setlength{\abovecaptionskip}{0cm}
	\centering
	\includegraphics[width=\linewidth]{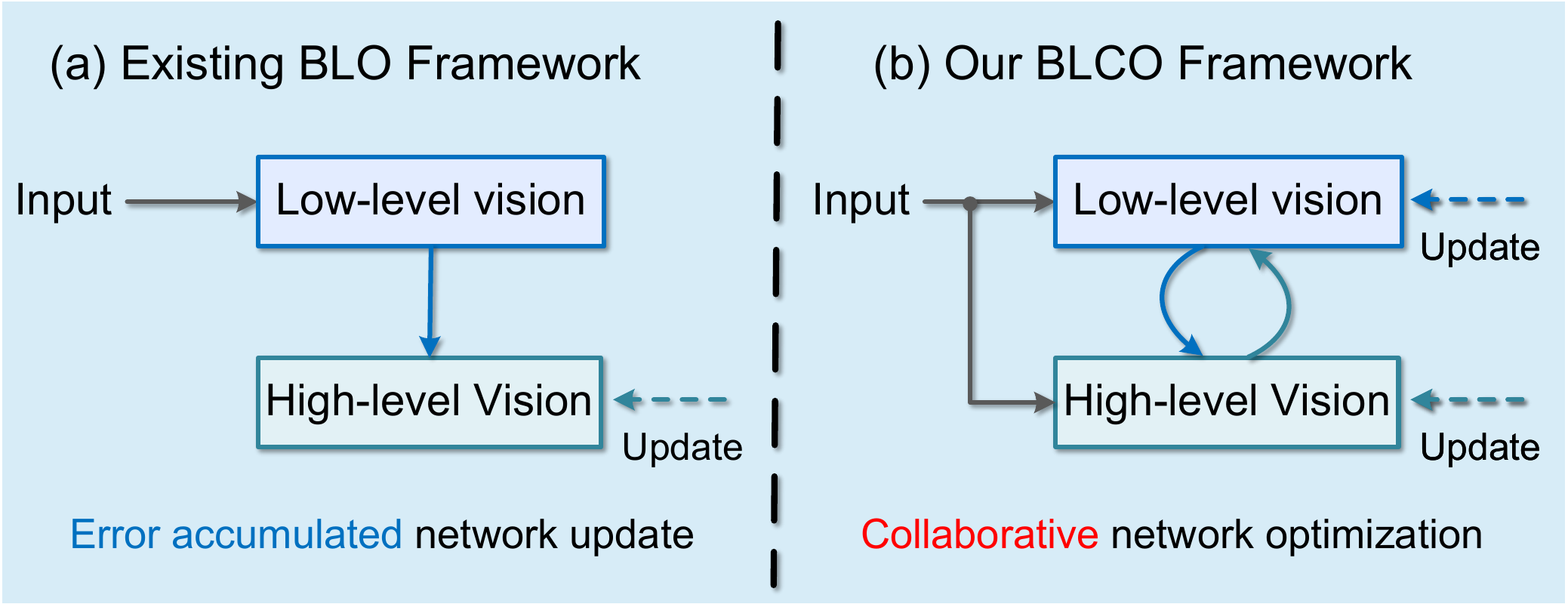 }
	\caption{Comparison between the existing bi-level optimization (BLO) framework and our bi-level cooperative optimization (BLCO) framework.}
	\label{fig:discussion}
\end{figure}

\noindent\textbf{Bi-level Optimization}. We generalize the core principle of RUN++—the joint optimization of segmentation and restoration—into a broader bi-level optimization framework. This framework aims to create a unified strategy that synergistically combines low-level and high-level vision tasks. By doing so, it is designed to enhance the robustness of high-level applications like segmentation and detection, particularly in challenging scenarios involving visual degradation (\textit{e.g.}, low light, low resolution, or haze), through collaborative support from low-level vision processes.

\begin{figure}[t]
\setlength{\abovecaptionskip}{0cm}
	\centering
	\includegraphics[width=\linewidth]{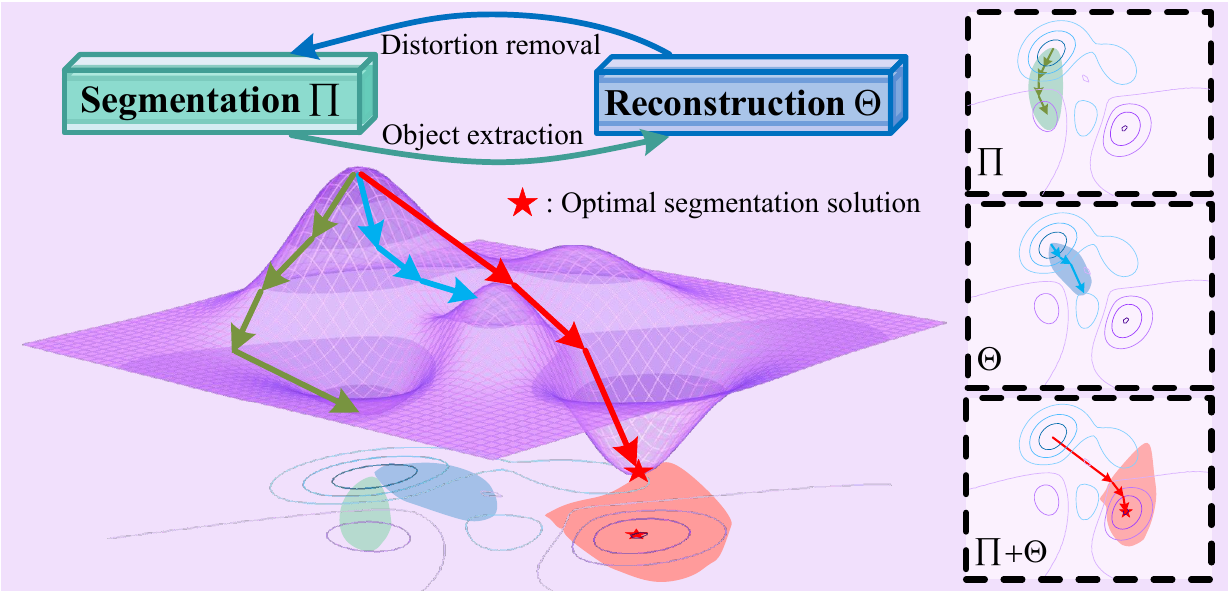}
	\caption{Conceptual illustration of the joint optimization strategy brought by our BLCO framework, where we take image segmentation (high-level vision) and image reconstruction (low-level vision) as an example. This facilitates the network's progression toward a more robust and accurate solution.
    }
	\label{fig:GradientOptimization1}
\end{figure}

\begin{table*}[tbp!]
		\setlength{\abovecaptionskip}{0cm} 
		\setlength{\belowcaptionskip}{-0.2cm}
		\centering
            \caption{Results on camouflaged object detection. SegMaR-1/-4 are SegMaR with one or four stages. 
            The best results are marked in \textbf{bold}. For the ResNet50 backbone in the common setting, the best two results are in {\color[HTML]{FF0000} \textbf{red}} and {\color[HTML]{00B0F0} \textbf{blue}} fonts.} \label{table:CODQuanti}
            \vspace{1mm}
		\resizebox{2.07\columnwidth}{!}{
			\setlength{\tabcolsep}{1.4mm}
			\begin{tabular}{l|c|cccc|cccc|cccc|cccc} 
				\toprule
				\multicolumn{1}{c|}{}                                        & \multicolumn{1}{c|}{}                           & \multicolumn{4}{c|}{\textit{CHAMELEON} }                                                                                                                                         & \multicolumn{4}{c|}{\textit{CAMO} }                                                                                                                                             & \multicolumn{4}{c|}{\textit{COD10K} }                                                                                                                                          & \multicolumn{4}{c}{\textit{NC4K} }                                                                                                                        \\ \cline{3-18} 
				\multicolumn{1}{l|}{\multirow{-2}{*}{Methods}} & \multicolumn{1}{c|}{\multirow{-2}{*}{Backbones}} & {\cellcolor{gray!40}$M$~$\downarrow$}                                  & {\cellcolor{gray!40}$F_\beta$~$\uparrow$}                               & {\cellcolor{gray!40}$E_\phi$~$\uparrow$}                               & \multicolumn{1}{c|}{\cellcolor{gray!40}$S_\alpha$~$\uparrow$}                                   & {\cellcolor{gray!40}$M$~$\downarrow$}                                  & {\cellcolor{gray!40}$F_\beta$~$\uparrow$}                               & {\cellcolor{gray!40}$E_\phi$~$\uparrow$}                               & \multicolumn{1}{c|}{\cellcolor{gray!40}$S_\alpha$~$\uparrow$}                                   & {\cellcolor{gray!40}$M$~$\downarrow$}                                  & {\cellcolor{gray!40}$F_\beta$~$\uparrow$}                               & {\cellcolor{gray!40}$E_\phi$~$\uparrow$}                               & \multicolumn{1}{c|}{\cellcolor{gray!40}$S_\alpha$~$\uparrow$}                                   & {\cellcolor{gray!40}$M$~$\downarrow$}                                  & {\cellcolor{gray!40}$F_\beta$~$\uparrow$}                               & {\cellcolor{gray!40}$E_\phi$~$\uparrow$}                               & \multicolumn{1}{c}{\cellcolor{gray!40}$S_\alpha$~$\uparrow$}                                   \\ \midrule 
				\multicolumn{18}{c}{Common Setting: Single Input Scale and Single Stage}                                   \\ \midrule
				\multicolumn{1}{l|}{LSR~\cite{lv2021simultaneously}}                       & \multicolumn{1}{c|}{ResNet50}                   & 0.030                                 & 0.835                                 & 0.935                                 & \multicolumn{1}{c|}{0.890}                                 & 0.080                                 & 0.756                                 & 0.838                                 & \multicolumn{1}{c|}{0.787}                                 & 0.037                                 & 0.699                                 & 0.880                                 & \multicolumn{1}{c|}{0.804}                                 & 0.048 & 0.802 & 0.890                                 & 0.834                                 \\
                \multicolumn{1}{l|}{FEDER~\cite{He2023Camouflaged}} & \multicolumn{1}{c|}{ResNet50}  & { {0.028}} & { {0.850}} & 0.944 & \multicolumn{1}{c|}{0.892} & {{0.070}} & {0.775} & 0.870 & \multicolumn{1}{c|}{0.802} & 0.032 & 0.715 & 0.892 & \multicolumn{1}{c|}{0.810} & {{0.046}} & {{0.808}} & {{0.900}} & {{0.842}} \\
                \multicolumn{1}{l|}{FGANet~\cite{zhaiexploring}}                     & \multicolumn{1}{c|}{ResNet50}                   & 0.030                                 & 0.838                                 & { {0.945}}                                 & 0.891                                 & 0.070  & 0.769  & 0.865  & \multicolumn{1}{c|}{0.800}  & 0.032   & 0.708 & 0.894  & \multicolumn{1}{c|}{0.803}                                 & 0.047                                & 0.800                                 & 0.891                                 & 0.837                                 \\
                \multicolumn{1}{l|}{FocusDiff~\cite{zhao2025focusdiffuser}} & \multicolumn{1}{c|}{ResNet50} & {{0.028}} & 0.843 & 0.938 & 0.890  & {\color[HTML]{00B0F0} \textbf{0.069}} & 0.772 & {\color[HTML]{00B0F0} \textbf{0.883}} & { {0.812}} & {{0.031}} & {{0.730}} & {{0.897}} & 0.820 & {{0.044}} & {{0.810}} & {{0.902}} &{{0.850}}    \\
                \multicolumn{1}{l|}{FSEL~\cite{sun2025frequency}} & \multicolumn{1}{c|}{ResNet50} & 0.029 & 0.847 & 0.941 & {{0.893}}  & {\color[HTML]{00B0F0} \textbf{0.069}} & {{0.779}} & { {0.881}} & {\color[HTML]{FF0000} \textbf{0.816}} & 0.032 & 0.722 & 0.891 & {{0.822}} & 0.045 & 0.807 & 0.901 & 0.847    \\ 
                \multicolumn{1}{l|}{RUN~\cite{he2025run}}  & \multicolumn{1}{c|}{ResNet50} & {\color[HTML]{00B0F0} \textbf{0.027}} & {\color[HTML]{00B0F0} \textbf{0.855}} & {\color[HTML]{00B0F0} \textbf{0.952}} & {\color[HTML]{00B0F0} \textbf{0.895}} & 0.070 & {\color[HTML]{00B0F0} \textbf{0.781}} & 0.868 & 0.806 & {\color[HTML]{00B0F0} \textbf{0.030}} & {\color[HTML]{00B0F0} \textbf{0.747}} & {\color[HTML]{00B0F0} \textbf{0.903}} & {\color[HTML]{00B0F0} \textbf{0.827}} & {\color[HTML]{00B0F0} \textbf{0.042}} & {\color[HTML]{00B0F0} \textbf{0.824}} & {\color[HTML]{00B0F0} \textbf{0.908}} & {\color[HTML]{00B0F0} \textbf{0.851}}     \\
                \rowcolor{c2!20} RUN++ (Ours) & ResNet50 &\color[HTML]{FF0000} \textbf{0.026} &\color[HTML]{FF0000} \textbf{0.870}  &\color[HTML]{FF0000} \textbf{0.956} &\color[HTML]{FF0000} \textbf{0.907}  & \color[HTML]{FF0000} \textbf{0.068} &\color[HTML]{FF0000} \textbf{0.787} &\color[HTML]{FF0000} \textbf{0.885} & \color[HTML]{FF0000}\textbf{0.816} & \color[HTML]{FF0000} \textbf{0.028} &\color[HTML]{FF0000} \textbf{0.758} & \color[HTML]{FF0000} \textbf{0.913} & \color[HTML]{FF0000} \textbf{0.832} & \color[HTML]{FF0000}\textbf{0.040} & \color[HTML]{FF0000}\textbf{0.829} & \color[HTML]{FF0000}\textbf{0.912} & \color[HTML]{FF0000}\textbf{0.853} \\
                \midrule
				\multicolumn{1}{l|}{BSA-Net~\cite{zhu2022can}}            & \multicolumn{1}{c|}{Res2Net50}                  & 0.027                                 & 0.851                                 & 0.946                                 & \multicolumn{1}{c|}{0.895}                                 & 0.079                                 & 0.768                                 & 0.851                                 & \multicolumn{1}{c|}{0.796}                                 & 0.034                                 & 0.723                                 & 0.891                                 & \multicolumn{1}{c|}{0.818}                                 & 0.048                                 & 0.805                                 & 0.897                                 & 0.841                                 \\
                \multicolumn{1}{l|}{FEDER~\cite{He2023Camouflaged} }                       & \multicolumn{1}{c|}{Res2Net50}   & 0.026 & 0.856  & 0.947  & 0.903  & {0.066}  & 0.807  & 0.897  & 0.836  & 0.029  & 0.748  & 0.911  & 0.844  & 0.042  & 0.824  & 0.913  & {0.862}               \\  
               \multicolumn{1}{l|}{RUN~\cite{he2025run}}  & \multicolumn{1}{c|}{Res2Net50} & {0.024} & {0.879} & {0.956} & {0.907} & {0.066} & {0.815} & {0.905} & {0.843}  & {0.028} & {0.764} & {0.914} & {0.849} & {0.041} & {0.830} & {0.917} &0.859                \\
                \rowcolor{c2!20}  RUN++ (Ours) & Res2Net50 & \textbf{0.023}&\textbf{0.883} &\textbf{0.965} &\textbf{0.922} & \textbf{0.065} & \textbf{0.820} & \textbf{0.912} & \textbf{0.845} & \textbf{0.026} & \textbf{0.782} & \textbf{0.927} & \textbf{0.855} & \textbf{0.039} & \textbf{0.837} & \textbf{0.927} & \textbf{0.865}   \\ 
                \midrule
                \multicolumn{1}{l|}{CamoFocus~\cite{khan2024camofocus} }               & \multicolumn{1}{c|}{PVT V2} & 0.023 & 0.869 & 0.953 & 0.906 & {0.044} & {0.861} & 0.924 &0.870 & 0.022 & {0.818} & 0.931 & 0.868 & 0.031 & 0.862 &0.932 &0.886 \\
                CamoDiff~\cite{sun2025conditional} & PVT V2 & 0.022 & 0.868 & 0.952 & 0.908 & 0.042 & 0.853 & 0.936 & 0.878 & 0.019 & 0.815 & 0.943 & 0.883 & 0.028 & 0.858 & 0.942 & 0.895 \\
                \multicolumn{1}{l|}{RUN~\cite{he2025run}} &  \multicolumn{1}{c|}{PVT V2} & {0.021} & {0.877} & {0.958} & {0.916} & 0.045 &  {0.861} &  {0.934} & {0.877} & {0.021} &  0.810 & {0.941} &  {0.878} & {0.030} & {0.868} & {0.940} & {0.892} \\ 
                \rowcolor{c2!20}  RUN++ (Ours) & PVT V2 & \textbf{0.019} & \textbf{0.886} & \textbf{0.959} & \textbf{0.920} & \textbf{0.041} & \textbf{0.869} & \textbf{0.944} & \textbf{0.883} & \textbf{0.019} & \textbf{0.836} & \textbf{0.949} & \textbf{0.885} & \textbf{0.029} & \textbf{0.876} & \textbf{0.941} & \textbf{0.895}
                \\
                \midrule
				\multicolumn{18}{c}{Other Setting: Multiple Input Scales (MIS)} \\ 
				 \midrule
				\multicolumn{1}{l|}{ZoomNet~\cite{pang2022zoom}}            & \multicolumn{1}{c|}{ResNet50}                   & 0.024                                 & 0.858                                 & 0.943                                 & \multicolumn{1}{c|}{0.902}                                 & 0.066                                 & 0.792                                 & 0.877                                 & \multicolumn{1}{c|}{0.820}                                 & 0.029                                 & 0.740                                 & 0.888                                 & \multicolumn{1}{c|}{0.838}                        & 0.043                                 & 0.814                                 & 0.896                                 & 0.853                                 \\
                 \multicolumn{1}{l|}{ZoomNext~\cite{pang2024zoomnext}} & \multicolumn{1}{c|}{ResNet50}  & {0.021} & 0.868 & 0.956 & 0.908 & 0.065 & 0.798 & 0.888 & 0.822 & 0.026 & 0.768 & 0.918 & {0.855} & {0.038} & 0.833 & 0.919 &0.869     \\
                RUN~\cite{he2025run} & \multicolumn{1}{c|}{ResNet50}  & {0.022} & {0.878} & { {0.967}} & {0.911} & {0.064} & {0.807} & { {0.902}} & {0.832} & {0.027} & {0.772} & {0.920} & {0.843} & {0.040} & {0.836} & {0.922} & {0.868}\\
                \rowcolor{c2!20}  RUN++ (Ours) & ResNet50 & \textbf{0.020} & \textbf{0.883} & \textbf{0.971} &  \textbf{0.914} & \textbf{0.063} & \textbf{0.812} & \textbf{0.904} & \textbf{0.836} & \textbf{0.025} & \textbf{0.777} & \textbf{0.926} & \textbf{0.856} & \textbf{0.037} & \textbf{0.841} & \textbf{0.932} & \textbf{0.872}    \\
                
                \midrule
				\multicolumn{18}{c}{Other Setting: Multiple Stages (MS)}  \\ \midrule
				\multicolumn{1}{l|}{SegMaR-4~\cite{jia2022segment}}   & \multicolumn{1}{c|}{ResNet50}                   & 0.025                        & 0.855                                 & 0.955                                 & \multicolumn{1}{c|}{0.906}                                 & 0.071                                 & 0.779                                 & 0.865                                 & \multicolumn{1}{c|}{0.815}                                 & 0.033                                 & 0.737                                 & 0.896                                 & \multicolumn{1}{c|}{0.833}                                 & 0.047                                 & 0.793                                 & 0.892                                 & 0.845                                 \\
                \multicolumn{1}{l|}{FEDER-4~\cite{He2023Camouflaged}}         & \multicolumn{1}{c|}{ResNet50}                  & 0.025 & 0.874  & 0.964  & 0.907  & 0.067  & 0.809  & 0.886  & 0.822  & 0.028  & 0.752  & 0.917  & 0.851  & 0.042  & 0.827  & 0.917  & 0.863         \\ 
                \multicolumn{1}{l|}{RUN-4~\cite{he2025run} } & \multicolumn{1}{c|}{ResNet50} & {0.024} & {0.889} & {0.968}&{0.913}& {0.066} & {0.815} & {0.893} & {0.829} &{0.027} &{0.769} &{0.926} &{0.857} & {0.041} & {0.833} & {0.925}& {0.870} \\ 
                \rowcolor{c2!20} RUN++-4 (Ours)  & ResNet50 & \textbf{0.023} & \textbf{0.891} & \textbf{0.974} & \textbf{0.915} & \textbf{0.065} & \textbf{0.823} & \textbf{0.901} & {\textbf{0.830}} & \textbf{0.026} & \textbf{0.778} & \textbf{0.928} & \textbf{0.858} & \textbf{0.039} & \textbf{0.837} & \textbf{0.933} & \textbf{0.875} \\ \midrule
                \multicolumn{18}{c}{Other Setting: High Resolution Segmentation (HRS)}  \\ \midrule
                BiRefNet~\cite{zheng2024bilateral} & SwinL & --- & --- & --- & --- & {\textbf{0.030}} & 0.895 & {0.954} &0.904 & 0.014 &0.881 & 0.960 &0.913 &0.023 & 0.925 &0.953 &0.914 \\
                FEDER~\cite{He2023Camouflaged}  & SwinL & --- & --- & --- & --- & 0.034 & 0.882 & 0.939 & 0.893 & 0.016 & 0.867 & 0.942 & 0.901 & 0.026 & 0.917 & 0.938 & 0.902 \\
                RUN~\cite{he2025run}  & SwinL & --- & --- & --- & --- & 0.032 & 0.896 & 0.946 & 0.900 & 0.014 & 0.879 & 0.957 & {0.915} & 0.024 & 0.928 & 0.955 & 0.910  \\
                \rowcolor{c2!20} RUN++~(Ours)  & SwinL & --- & --- & --- & --- & \textbf{0.030} & \textbf{0.899} & \textbf{0.957} & \textbf{0.906}& \textbf{0.012} & \textbf{0.887} & \textbf{0.968} & \textbf{0.916} & \textbf{0.022} & \textbf{0.937} & \textbf{0.960} & \textbf{0.918}  \\
                \bottomrule            
		\end{tabular}}
		\vspace{-1mm}
	\end{table*}
The motivation for this bi-level approach stems from a critical observation: high-fidelity image restoration (a low-level task) does not guarantee improved performance on a subsequent high-level task. As illustrated in~\cref{fig:discussion} (a), existing methods~\cite{liu2022target,he2023HQG} typically employ a sequential, or tandem, approach. The limitations of this pipeline are twofold. First, it risks propagating artifacts from the restoration output, which can degrade the performance of the high-level task. Second, and more critically, it overlooks the potential for reciprocal feedback, where guidance from the high-level task can, in turn, enhance the low-level restoration process.

For example, segmentation errors often occur at object boundaries, which contain crucial structural information that is also inherently challenging for image restoration. In this context, guidance from the segmentation network can direct the restoration network to focus specifically on these challenging boundary regions, leading to restorations with superior structural preservation and textural detail.

Thus, a framework that enables the co-optimization of low-level and high-level tasks can achieve superior performance in both the visual fidelity of the restored image and the robustness of the downstream task. Drawing inspiration from Stackelberg game theory, which is foundational to bi-level optimization~\cite{ochs2015bilevel}, we formulate our proposed strategy, termed Bi-Level Collaborative Optimization (BLCO), as follows:
\begin{equation}
				\underset{\bm{\omega}_l,\bm{\omega}_h}{\min} \ \mathcal{L}^{l}\left(\Theta\left({\mathbf{I}},f^\Pi;\bm{\omega}_l\right)\right)+\mathcal{L}^{h}\left(\Pi\left({\mathbf{I}},f^\Theta;\bm{\omega}_h\right)\right), 
		\end{equation}
where $\mathcal{L}^{l}$ and $\mathcal{L}^{h}$ are the objective functions for the low-level and high-level vision tasks. The network $\Theta\left(\bigcdot;\bm{\omega}_l\right)$ and $\Pi\left(\bigcdot;\bm{\omega}_h\right)$ are parameterized by the trainable weights $\bm{\omega}_l$ and $\bm{\omega}_h$. $f^\Pi$ and $f^\Theta$ are the guidance information (e.g., intermediate features or final outputs) exchanged between the two tasks.

As shown in~\cref{fig:GradientOptimization1}, BLCO motivates further exploration of collaborative optimization strategies to enhance the resilience of high-level vision algorithms to environmental degradation and imaging interference. Simultaneously, it advances low-level vision algorithms by enabling the integration of high-level semantic guidance. This dual enhancement fosters the development of more robust and reliable vision systems capable of performing accurately in complex, real-world environments.

\begin{table*}[ht]
\setlength{\abovecaptionskip}{0cm} 
\setlength{\belowcaptionskip}{-0.2cm}
\centering
\caption{Efficiency comparison with existing cutting-edge methods on COD, including three commonly-used backbones. } \label{table:efficiency}
\vspace{1mm}
\resizebox{2.07\columnwidth}{!}{
\setlength{\tabcolsep}{1.6mm}
\begin{tabular}{l|ccc|ccc|ccc}
\toprule
\multirow{2}{*}{Efficiency} & \multicolumn{3}{c|}{ResNet50} & \multicolumn{3}{c|}{Res2Net50} & \multicolumn{3}{c}{PVT V2} \\ \cline{2-10}
                            & FocusDiff~\cite{zhao2025focusdiffuser} & RUN~\cite{he2025run}    & \cellcolor{c2!20}  RUN++ (Ours)  & FEDER~\cite{He2023Camouflaged}    & RUN~\cite{he2025run}      &\cellcolor{c2!20}  RUN++ (Ours)    & CamoFocus~\cite{khan2024camofocus}  & RUN~\cite{he2025run}   &\cellcolor{c2!20}  RUN++ (Ours)  \\ \midrule
Parameters (M) $\downarrow$             & 166.17   & 30.41  & \cellcolor{c2!20}  36.52   & 45.92    & 30.57     & \cellcolor{c2!20}  36.69    & 68.85     & 65.17 &   \cellcolor{c2!20}  68.64    \\
FLOPs (G) $\downarrow$                  & 7618.49  &    43.36  & \cellcolor{c2!20}  47.16    & 50.03    & 45.73    & \cellcolor{c2!20}  48.90        & 91.35       & 61.83 & \cellcolor{c2!20}   68.33    \\
FPS $\uparrow$                        & 0.23     & 22.75  & \cellcolor{c2!20} 15.51      & 14.02    & 20.26    & \cellcolor{c2!20}   15.19      & 9.63       & 15.82 & \cellcolor{c2!20}    11.73  \\ \bottomrule
\end{tabular}}\vspace{-3mm}
\end{table*}

\begin{table}[ht]
\setlength{\abovecaptionskip}{0cm} 
		\setlength{\belowcaptionskip}{-0.2cm}
\centering
\caption{Results on polyp image segmentation.
        } \label{table:PISQuanti}
	\resizebox{1\columnwidth}{!}{
		\setlength{\tabcolsep}{1.2mm}
	\begin{tabular}{l|ccc|ccc}
		\toprule 
		\multirow{2}{*}{Methods} & \multicolumn{3}{c|}{\textit{CVC-ColonDB} }  & \multicolumn{3}{c}{\textit{ETIS} } 
        \\ \cline{2-7} 
		& \multicolumn{1}{c}{\cellcolor{gray!40}mDice~$\uparrow$} & \multicolumn{1}{c}{\cellcolor{gray!40}mIoU~$\uparrow$} & \multicolumn{1}{c|}{\cellcolor{gray!40}$S_\alpha$~$\uparrow$} & \multicolumn{1}{c}{\cellcolor{gray!40}mDice~$\uparrow$} & \multicolumn{1}{c}{\cellcolor{gray!40}mIoU~$\uparrow$} & \multicolumn{1}{c}{\cellcolor{gray!40}$S_\alpha$~$\uparrow$} \\ \midrule
		PraNet~\cite{fan2020pranet} & 0.709 & 0.640  & 0.819  &  0.628 & 0.567 & 0.794 \\
        CASCADE~\cite{rahman2023medical} & 0.809 & 0.731 & 0.867 & 0.781 & {{0.706}} & 0.853  \\
        PolypPVT~\cite{dong2023polyp}  & 0.808 &0.727 & 0.865 & {{0.787}} & {{0.706}} & {{0.871}} \\
        CoInNet~\cite{jain2023coinnet} & 0.797 &0.729 &{{0.875}} & 0.759 & 0.690 & 0.859 \\
        LSSNet~\cite{wang2024lssnet} & {{0.820}} & {{0.741}} & 0.867 & 0.779 & 0.701 & 0.867 \\
        RUN~\cite{he2025run}  & {\color[HTML]{00B0F0} \textbf{0.822}} & {\color[HTML]{00B0F0} \textbf{0.742}} & {\color[HTML]{00B0F0} \textbf{0.880}} & {\color[HTML]{00B0F0} \textbf{0.788}} &{\color[HTML]{00B0F0} \textbf{0.709}} & {\color[HTML]{00B0F0} \textbf{0.878}}   \\
        \rowcolor{c2!20} RUN++ (Ours) &\color[HTML]{FF0000} \textbf{0.826} &\color[HTML]{FF0000} \textbf{0.749} &\color[HTML]{FF0000} \textbf{0.886} & \color[HTML]{FF0000} \textbf{0.790} & \color[HTML]{FF0000} \textbf{0.714} & \color[HTML]{FF0000} \textbf{0.880}  \\
	 \bottomrule                      
	\end{tabular}}\label{table:polyp}
		\vspace{-0.1cm}
	\end{table}

\section{Experiments}\label{Sec:Experiment}
\subsection{Experimental Setup}\label{Sec:ExperimentSet}
RUN++ is implemented using PyTorch and trained on two RTX 4090 GPUs. Following common practice~\cite{fan2020camouflaged,He2023Camouflaged}, we integrate deep features from a shared encoder. For training and inference, all images are resized to $352 \times 352$.
We use Adam with momentum parameters of $\beta_1=0.9$ and $\beta_2=0.999$. A batch size of 36 is selected with an initial learning rate of $1\times10^{-4}$, which is decayed by a factor of 10 every 80 epochs. The number of unfolding stages, $K$, is set to 3. For the diffusion process within the FINE module, we employed a cosine noise schedule with $T=1000$ timesteps during training. For efficient inference, we utilized DDIM sampling with 5 timesteps and only set FINE in the last stage. All extra parameters inherited from the original optimization model are treated as learnable variables with random initialization.

\begin{table}[ht]
\setlength{\abovecaptionskip}{0cm} 
		\setlength{\belowcaptionskip}{-0.2cm}
	\centering
        \caption{ Results on medical tubular object segmentation.
        } \label{table:MTOSQuanti}
	\resizebox{1\columnwidth}{!}{
		\setlength{\tabcolsep}{1.2mm}
	\begin{tabular}{l|ccc|ccc}
		\toprule 
		\multirow{2}{*}{Methods} & \multicolumn{3}{c|}{\textit{DRIVE} }  & \multicolumn{3}{c}{\textit{CORN} } 
        \\ \cline{2-7} 
		& \multicolumn{1}{c}{\cellcolor{gray!40}mDice~$\uparrow$} & {\cellcolor{gray!40}AUC~$\uparrow$}                               & \multicolumn{1}{c|}{\cellcolor{gray!40}SEN~$\uparrow$} & \multicolumn{1}{c}{\cellcolor{gray!40}mDice~$\uparrow$} & {\cellcolor{gray!40}AUC~$\uparrow$}                               & \multicolumn{1}{c|}{\cellcolor{gray!40}SEN~$\uparrow$}  \\ \midrule
		CS2-Net~\cite{mou2021cs2}  & 0.795                     & {{0.983}}                   & 0.822                    & 0.607                     & 0.960                   & 0.817       \\
        DSCNet~\cite{qi2023dynamic}   & 0.805                     & 0.955                   & 0.830                    & 0.618                     & {\color[HTML]{00B0F0} {\textbf{0.964}}}                   & 0.856     \\
        SGAT~\cite{lin2023stimulus}   & {{0.806}}                     & 0.953                   & {{0.832}}                    & 0.639                     & 0.961                   & 0.853      \\
        TAU~\cite{gupta2024topology}    & 0.798                     & 0.977                   & 0.825                    & 0.643                     & 0.949                   & {{0.859}}      \\
        FFM~\cite{huang2025representing}    & 0.791                     & 0.972                   & 0.830                    & {{0.647}}                     & 0.952                   & 0.835              \\
        RUN~\cite{he2025run} & {\color[HTML]{00B0F0} \textbf{0.812}}                     & {\color[HTML]{00B0F0} \textbf{0.985}}                   & {\color[HTML]{00B0F0} \textbf{0.845}}                    & {\color[HTML]{00B0F0} \textbf{0.652}}                      & {{0.962}}                   & {\color[HTML]{00B0F0} \textbf{0.870}}           \\
        \rowcolor{c2!20} RUN++ (Ours) & \color[HTML]{FF0000} \textbf{0.817} & \color[HTML]{FF0000}\textbf{0.990} & \color[HTML]{FF0000}\textbf{0.850} & \color[HTML]{FF0000}\textbf{0.662} & \color[HTML]{FF0000}\textbf{0.971} &\color[HTML]{FF0000} \textbf{0.873} \\
	 \bottomrule                      
	\end{tabular}}\label{table:vessel}
		\end{table}

\subsection{Comparisons in CVP Tasks}
We benchmarked our method against SOTA competitors across several CVP tasks. To ensure fair and reproducible comparisons, all methods were evaluated using publicly available toolkits. It should be noted that while the COD domain offers a rich set of open-sourced methods for extensive quantitative analysis, other specialized CVP sub-tasks have a more limited number of public baselines, limiting quantitative analysis. 	

\begin{figure*}[t]
\setlength{\abovecaptionskip}{0cm}
	\centering
	\includegraphics[width=\linewidth]{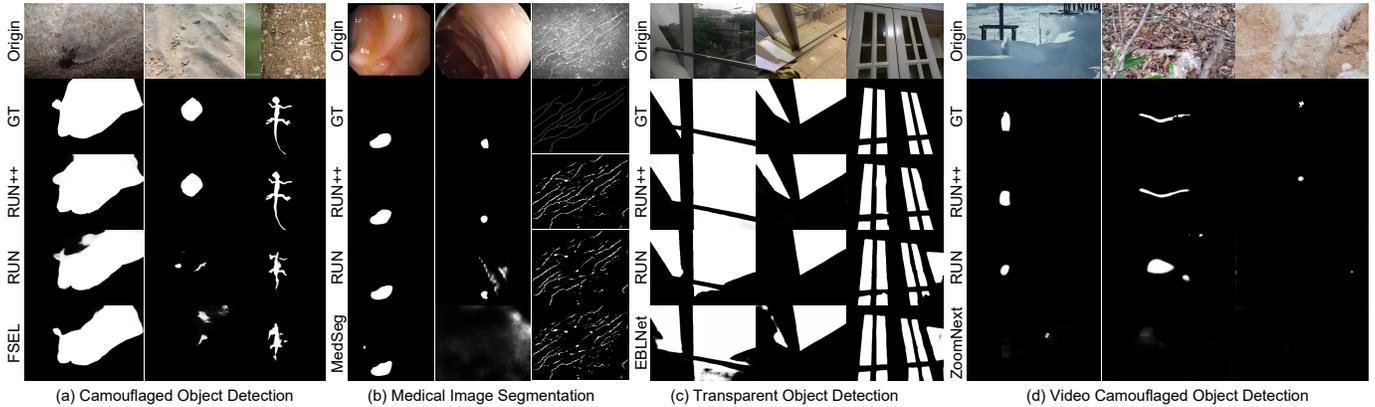}
	\caption{Visual comparison on COD, MIS, TOD, and VCOD tasks. In (b), MedSeg corresponds to PolypPVT~\cite{dong2023polyp} and DSCNet~\cite{qi2023dynamic} for the polyp image segmentation and medical tubular object segmentation tasks, respectively.}
	\label{fig:MedQuali1}
	\vspace{-3mm}
\end{figure*}

{\noindent \textbf{Camouflaged object detection}. In COD, we perform experiments using four benchmark datasets: \textit{CHAMELEON}~\cite{skurowski2018animal}, \textit{CAMO}~\cite{le2019anabranch}, \textit{COD10K}~\cite{fan2021concealed}, and \textit{NC4K}~\cite{lv2021simultaneously}. 
Our training set comprised 1,000 images from CAMO and 3,040 images from COD10K. The test set was formed using the remaining images from these two datasets, supplemented by the complete CHAMELEON and NC4K datasets.
Performance was assessed using four standard evaluation metrics: mean absolute error ($M$), adaptive F-measure ($F_\beta$)~\cite{margolin2014evaluate}, mean E-measure ($E_\phi$)~\cite{fan2021cognitive}, and structure measure ($S_\alpha$)~\cite{fan2017structure}. Optimal results are indicated by a lower value of $M$ and higher values for $F_\beta$, $E_\phi$, and $S_\alpha$. 
As detailed in ~\cref{table:CODQuanti}, our method achieves SOTA performance in all four experimental settings. Within the common setting, our approach surpasses the second-best methods across three distinct backbones: ResNet50~\cite{he2016deep}, Res2Net50~\cite{gao2019res2net}, and PVT V2~\cite{wang2022pvt}, representing average performance gains of $1.87\%$, $1.84\%$, and $2.61\%$ over the second-best method (RUN~\cite{he2025run}), respectively, calculated across the four datasets. 
Furthermore, in the specialized MIS, MS, and HRS settings, our framework adheres to the respective evaluation protocols of ZoomNet~\cite{pang2022zoom}, SegMaR~\cite{jia2022segment}, and BiRefNet~\cite{zheng2024bilateral} (only evaluating performance in three datasets except \textit{CHAMELEON}), consistently delivering improved results over these existing methods. RUN++ generally surpasses the top-performing methods by $1.48\%$ (\textit{vs}. ZoomNext~\cite{pang2024zoomnext} in MIS), $1.30\%$ (\textit{vs}. RUN~\cite{he2025run} in MS), and $2.00\%$ (\textit{vs}. BiRefNet~\cite{zheng2024bilateral} in HRS). \begin{table}[tbp!]
	\setlength{\abovecaptionskip}{0cm} 
		\setlength{\belowcaptionskip}{-0.2cm}
	\centering
        \caption{ Results on transparent object detection. 
        } \label{table:TODQuanti}
	\resizebox{\columnwidth}{!}{
		\setlength{\tabcolsep}{1mm}
	\begin{tabular}{l|ccc|ccc}\toprule 

		\multicolumn{1}{l|}{}                          & \multicolumn{3}{c|}{\textit{GDD} } & \multicolumn{3}{c}{\textit{GSD} }  \\ \cline{2-7} 
		\multicolumn{1}{l|}{\multirow{-2}{*}{Methods}} & \cellcolor{gray!40}mIoU~$\uparrow$&\cellcolor{gray!40}$F_\beta^{max}$~$\uparrow$&\cellcolor{gray!40} $M$~$\downarrow$& \cellcolor{gray!40}mIoU~$\uparrow$&\cellcolor{gray!40}$F_\beta^{max}$~$\uparrow$&\cellcolor{gray!40} $M$~$\downarrow$ \\ \midrule
		GDNet~\cite{mei2020don}                                          & 0.876                                 & 0.937                                 & 0.063                                                              & 0.790                                 & 0.869                                 & 0.069    \\
		EBLNet~\cite{he2021enhanced} & 0.870                                 & 0.922                                 & 0.064                                  & 0.817                                 & 0.878                                 & 0.059 \\
        RFENet~\cite{fan2023rfenet} & 0.886 & 0.938 & 0.057 & {0.865} & {{0.931}} & {{0.048}} \\
        IEBAF~\cite{han2023internal} & 0.887 & {{0.944}} & 0.056  &  0.861 & 0.926 & 0.049  \\
        GhostingNet~\cite{yan2024ghostingnet} & {{0.893}}  & 0.943 & {{0.054}} & 0.838 & 0.904 & 0.055   \\
	RUN~\cite{he2025run}  & {\color[HTML]{00B0F0} \textbf{0.895}} & {\color[HTML]{00B0F0} \textbf{0.952}} & {\color[HTML]{00B0F0} \textbf{0.051}} & {\color[HTML]{00B0F0} \textbf{0.866}} & {\color[HTML]{00B0F0} \textbf{0.938}} & {\color[HTML]{00B0F0} \textbf{0.043}}   \\ 
    \rowcolor{c2!20} 	RUN++ (Ours)  &\color[HTML]{FF0000} \textbf{0.899} & \color[HTML]{FF0000}\textbf{0.959} &\color[HTML]{FF0000} \textbf{0.050} &\color[HTML]{FF0000} \textbf{0.870} &\color[HTML]{FF0000} \textbf{0.946} & \color[HTML]{FF0000}\textbf{0.041} \\
    \bottomrule  \end{tabular}}
\end{table}
Qualitatively, as illustrated in ~\cref{fig:MedQuali1}, our approach generates segmentation maps that are significantly more complete and precise. We attribute this improvement to our novel, jointly reversible modeling at both the mask and RGB levels.
In addition to performance, we analyze the efficiency of our method in Table~\ref{table:efficiency}. The results demonstrate that the components introduced in RUN++ add only a modest computational overhead compared to the original RUN architecture~\cite{he2025run}. Specifically, the diffusion-based refiner and multi-stage design achieve significant performance gains without imposing a prohibitive increase in computational cost, affirming the overall efficiency of our framework.
}

{\noindent \textbf{Medical concealed visual perception}. We evaluated our method on two distinct medical CVP tasks, including polyp image segmentation (PIS) and medical tubular object segmentation (MTOS). For PIS, experiments were conducted on the \textit{CVC-ColonDB}~\cite{tajbakhsh2015automated} and \textit{ETIS}~\cite{silva2014toward} datasets. For MTOS, we utilized the \textit{DRIVE}\footnote{http://www.isi.uu.nl/Research/Databases/DRIVE/} and \textit{CORN}~\cite{ma2021structure} datasets.
The experimental protocol for PIS followed the setup established by LSSNet~\cite{wang2024lssnet}. Performance was evaluated using three metrics: mean Dice (mDice), mean Intersection over Union (mIoU), and $S_\alpha$. 
For MTOS, the training and test sets were partitioned according to RUN~\cite{he2025run}, and performance was measured using mDice, area under the ROC curve (AUC), and sensitivity (SEN). For all metrics across both tasks, higher values denote superior performance.
In line with recent state-of-the-art methods that commonly employ Transformer-based encoders, we adopted PVT V2 as our default encoder architecture. As presented in~\cref{table:PISQuanti,table:MTOSQuanti}, our method achieves leading performance across both segmentation tasks. Furthermore, the qualitative results in~\cref{fig:MedQuali1} visually confirm the effectiveness of our approach in accurately segmenting challenging structures, such as small polyps and fine-grained nerves. }
\begin{table}[t]
	\centering
	\setlength{\abovecaptionskip}{0cm} 
		\setlength{\belowcaptionskip}{-0.2cm}
	\caption{Results on concealed defect detection.  
    } \label{table:CDDQuanti}
	\resizebox{1\columnwidth}{!}{
		\setlength{\tabcolsep}{1.43mm}
		\begin{tabular}{l|cccccccc}
        \toprule 
Methods    & {\cellcolor{gray!40}$S_\alpha$~$\uparrow$} & {\cellcolor{gray!40}$M$~$\downarrow$} & {\cellcolor{gray!40}$E_\phi$~$\uparrow$} & {\cellcolor{gray!40}$E_\phi^{max}$~$\uparrow$} & {\cellcolor{gray!40}$F_\beta$~$\uparrow$} & {\cellcolor{gray!40}$F_\beta^{mean}$~$\uparrow$} \\ \midrule
SINet V2~\cite{fan2021concealed}   & 0.551 & 0.102 & 0.567  & {{0.597}}  & 0.223  & 0.248  \\
HitNet~\cite{hu2022high}     & 0.563 & 0.118 & 0.564  & 0.570  & {{0.298}}  & 0.298 \\
CamoFormer~\cite{yin2024camoformer} & {{0.589}} & {{0.100}} & {{0.588}}  & 0.596  & {\color[HTML]{00B0F0} \textbf{0.330}}  & {\color[HTML]{00B0F0} \textbf{0.329}}  \\
OAFormer~\cite{yang2023oaformer}   & 0.541 & 0.121 & 0.535  & 0.591  & 0.216  & 0.239  \\
FEDER~\cite{He2023Camouflaged} & 0.538 & 0.070 & 0.586 & 0.602 & 0.288 & 0.293 \\
RUN~\cite{he2025run} & {\color[HTML]{00B0F0} \textbf{0.590}} &  {\color[HTML]{00B0F0} \textbf{0.068}} & {\color[HTML]{00B0F0} \textbf{0.595}} & {\color[HTML]{00B0F0} \textbf{0.611}}  & {{0.298}}  & {{0.299}} \\
\rowcolor{c2!20} RUN++ (Ours) &\color[HTML]{FF0000} \textbf{0.602} & \color[HTML]{FF0000}\textbf{0.065} & \color[HTML]{FF0000}\textbf{0.610} & \color[HTML]{FF0000}\textbf{0.628} & \color[HTML]{FF0000}\textbf{0.332} & \color[HTML]{FF0000}\textbf{0.334}  \\
\bottomrule
\end{tabular}}
\end{table}

\begin{table*}[ht]
\setlength{\abovecaptionskip}{0cm} 
		\setlength{\belowcaptionskip}{-0.2cm}
	\centering
	\caption{Results on label-deficient concealed visual perception, where we take COD as an example and conduct experiments with weak supervision (scribble supervision) and semi-supervision (1/16 labeled training data). To accommodate the label-deficient settings, we integrate RUN~\cite{he2025run} and our RUN++ with the SEE framework~\cite{he2025segment} by replacing its own segmenter. 
        }
	\resizebox{2\columnwidth}{!}{
		\setlength{\tabcolsep}{1.3mm}
		\begin{tabular}{l|c|cccc|cccc|cccc|cccc}
			\toprule
			\multicolumn{1}{l|}{} & \multicolumn{1}{c|}{} & \multicolumn{4}{c|}{\textit{CHAMELEON} } & \multicolumn{4}{c|}{\textit{CAMO}} & \multicolumn{4}{c|}{\textit{COD10K}} & \multicolumn{4}{c}{\textit{NC4K}} \\ \cline{3-18}
			\multicolumn{1}{l|}{\multirow{-2}{*}{Methods}} & \multicolumn{1}{c|}{\multirow{-2}{*}{Pub.}} & {\cellcolor{gray!40}$M$~$\downarrow$} &{\cellcolor{gray!40}$F_\beta$~$\uparrow$} &{\cellcolor{gray!40}$E_\phi$~$\uparrow$} & \multicolumn{1}{c|}{\cellcolor{gray!40}$S_\alpha$~$\uparrow$}& {\cellcolor{gray!40}$M$~$\downarrow$} &{\cellcolor{gray!40}$F_\beta$~$\uparrow$} &{\cellcolor{gray!40}$E_\phi$~$\uparrow$} & \multicolumn{1}{c|}{\cellcolor{gray!40}$S_\alpha$~$\uparrow$}& {\cellcolor{gray!40}$M$~$\downarrow$} &{\cellcolor{gray!40}$F_\beta$~$\uparrow$} &{\cellcolor{gray!40}$E_\phi$~$\uparrow$} & \multicolumn{1}{c|}{\cellcolor{gray!40}$S_\alpha$~$\uparrow$}& {\cellcolor{gray!40}$M$~$\downarrow$} &{\cellcolor{gray!40}$F_\beta$~$\uparrow$} &{\cellcolor{gray!40}$E_\phi$~$\uparrow$} & \multicolumn{1}{c}{\cellcolor{gray!40}$S_\alpha$~$\uparrow$}\\ \midrule
			\multicolumn{18}{c}{Scribble Supervision} \\ \midrule
		 $\text{SAM-W}$~\cite{chen2023sam}                                         & \multicolumn{1}{c|}{ICCVW23}                              & 0.069                & 0.751                & 0.835                & 0.661                & 0.097                & 0.696                & 0.788                & 0.738                & 0.049                & 0.712                & 0.833                & 0.770                & 0.066                & 0.757                & 0.842                & 0.768                \\
			SCOD~\cite{he2022weakly}                                         & AAAI23                                             & 0.046                                 &\color[HTML]{00B0F0} \textbf{0.791}                                 & 0.897                                 & 0.818                                 & 0.092                                 & 0.709                                 & 0.815                                 & 0.735                                 & 0.049                                 & 0.637                                 & 0.832                                 & 0.733                                 & 0.064                                 & 0.751                                 & 0.853                                 & 0.779                                 \\
      GenSAM~\cite{hu2024relax}   & AAAI24 & 0.090 & 0.680 & 0.807 & 0.764 & 0.113 & 0.659 & 0.775 & 0.719 & 0.067 & 0.681 & 0.838 & 0.775 & 0.097 & 0.687 & 0.750 & 0.732 \\
WS-SAM~\cite{he2023weaklysupervised} & NIPS23 & 0.046                                 & 0.777                                 & 0.897                                 & {{0.824}} & 0.092                                 & {{0.742}} & 0.818                                 & {{0.759}} & {{0.038}} & {{0.719}} & {{0.878}} & {{0.803}} & {{0.052}} & {{0.802}} & {{0.886}} & {{0.829}} \\ 
SEE~\cite{he2025segment} & \multicolumn{1}{c|}{TPAMI25}& {\color[HTML]{00B0F0} \textbf{0.044}} & 0.785                                 & {\color[HTML]{00B0F0} \textbf{0.903}} & {\color[HTML]{00B0F0} \textbf{0.826}} & {\color[HTML]{00B0F0} \textbf{0.090}} & {{0.747}} & {\color[HTML]{00B0F0} \textbf{0.826}} & { {0.765}} & {\color[HTML]{00B0F0} \textbf{0.036}} & {\color[HTML]{00B0F0} \textbf{0.729}} & {\color[HTML]{00B0F0} \textbf{0.883}} & {{0.807}} & {{0.051}} & {\color[HTML]{00B0F0} \textbf{0.808}} & {\color[HTML]{00B0F0} \textbf{0.891}} & {\color[HTML]{00B0F0} \textbf{0.836}} \\
RUN~\cite{he2025run} & ICML25 & {\color[HTML]{00B0F0} \textbf{0.044}} & 0.783 & 0.898 & 0.823 & 0.091 &\color[HTML]{00B0F0} \textbf{0.750} & \color[HTML]{00B0F0} \textbf{0.826} & \color[HTML]{00B0F0} \textbf{0.767} &  \color[HTML]{00B0F0} \textbf{0.036} & 0.725 & 0.881 & {\color[HTML]{00B0F0} \textbf{0.808}} &\color[HTML]{00B0F0} \textbf{0.050} &  0.806 & 0.888 & 0.834     \\
\rowcolor{c2!20} RUN++ (Ours) & --- &\color[HTML]{FF0000} \textbf{0.042} &\color[HTML]{FF0000} \textbf{0.793} &\color[HTML]{FF0000} \textbf{0.910} &\color[HTML]{FF0000} \textbf{0.829} &\color[HTML]{FF0000} \textbf{0.088} &\color[HTML]{FF0000} \textbf{0.752} &\color[HTML]{FF0000} \textbf{0.831} &\color[HTML]{FF0000} \textbf{0.769} &\color[HTML]{FF0000} \textbf{0.035} &\color[HTML]{FF0000} \textbf{0.733} &\color[HTML]{FF0000} \textbf{0.889} &\color[HTML]{FF0000} \textbf{0.813} &\color[HTML]{FF0000} \textbf{0.049} &\color[HTML]{FF0000} \textbf{0.812} &\color[HTML]{FF0000} \textbf{0.895} &\color[HTML]{FF0000} \textbf{0.839}   \\
   \midrule
			\multicolumn{18}{c}{1/16 Labeled Training Data} \\ \midrule
SAM-S~\cite{chen2023sam}      &    ICCVW23    & 0.150                                 & 0.642                                 & 0.680                                 & 0.657                                 & 0.146                                 & 0.624                                 & 0.663                                 & 0.647                                 & 0.078                                 & {\color[HTML]{00B0F0} \textbf{0.682}} & 0.752                                 & 0.738                                 & 0.093                                 & 0.692                                 & 0.741                                 & 0.753                                 \\
PGCL~\cite{basak2023pseudo}       & CVPR23                & 0.057                                 & 0.752                                 & 0.850                                 & 0.801                                 & 0.116                                 & 0.682                                 & 0.782                                 & 0.722                                 & 0.061                                 & 0.637                                 & 0.779                                 & 0.728                                 & 0.071                                 & 0.719                                 & 0.809                                 & 0.789                                 \\
EPS~\cite{lee2023saliency}        & TPAMI23               & 0.049                                 & 0.763                                 & 0.843                                 & 0.828                                 & 0.103                                 & 0.697                                 & 0.796                                 & 0.735                                 & 0.056                                 & 0.646                                 & 0.787                                 & 0.736                                 & 0.066                                 & 0.737                                 & 0.833                                 & 0.801                                 \\
CoSOD~\cite{chakraborty2024unsupervised}      & WACV24                & 0.055                                 & 0.758                                 & 0.856                                 & 0.830                                 & 0.099                                 & 0.702                                 & 0.793                                 & 0.730                                 & 0.055                                 & 0.650                                 & 0.795                                 & 0.740                                 & 0.070                                 & 0.726                                 & 0.825                                 & 0.792                                 \\
SEE~\cite{he2025segment} & TPAMI25   & {\color[HTML]{00B0F0} \textbf{0.040}} & {\color[HTML]{00B0F0} \textbf{0.793}} & { {0.885}} & {\color[HTML]{00B0F0} \textbf{0.852}} & {{0.093}} & {{0.716}} & {{0.810}} & {\color[HTML]{00B0F0} \textbf{0.747}} & {{0.046}} & {{0.679}} & {\color[HTML]{00B0F0} \textbf{0.803}} & {{0.745}} & {{0.060}} & {{0.757}} & {{0.863}} & {\color[HTML]{00B0F0} \textbf{0.812}} \\ 
RUN~\cite{he2025run} & ICML25 & 0.041 & 0.790 &\color[HTML]{00B0F0} \textbf{0.890} & 0.847 & \color[HTML]{00B0F0} \textbf{0.092} & \color[HTML]{00B0F0} \textbf{0.723} &\color[HTML]{00B0F0} \textbf{0.816} & 0.746 & \color[HTML]{00B0F0} \textbf{0.045} & 0.676 & 0.802 & \color[HTML]{00B0F0} \textbf{0.748} &\color[HTML]{00B0F0} \textbf{0.059} & \color[HTML]{00B0F0} \textbf{0.766} & \color[HTML]{00B0F0} \textbf{0.866} & \color[HTML]{00B0F0} \textbf{0.812}  \\
\rowcolor{c2!20} RUN++ (Ours) & --- &\color[HTML]{FF0000} \textbf{0.039} &\color[HTML]{FF0000} \textbf{0.799}& \color[HTML]{FF0000} \textbf{0.897} & \color[HTML]{FF0000} \textbf{0.856} & \color[HTML]{FF0000} \textbf{0.090} & \color[HTML]{FF0000} \textbf{0.732} & \color[HTML]{FF0000} \textbf{0.828} & \color[HTML]{FF0000} \textbf{0.753} & \color[HTML]{FF0000} \textbf{0.043} & \color[HTML]{FF0000} \textbf{0.692} &\color[HTML]{FF0000} \textbf{0.818} & \color[HTML]{FF0000} \textbf{0.756}  & \color[HTML]{FF0000} \textbf{0.057}  & \color[HTML]{FF0000} \textbf{0.775}  & \color[HTML]{FF0000} \textbf{0.875} & \color[HTML]{FF0000} \textbf{0.818}   \\
\bottomrule
	\end{tabular}}
	\label{table:WCODQuanti}
	\vspace{-0.3cm}
\end{table*}
\begin{table*}[htp]
\begin{minipage}[c]{\textwidth}
\centering
\setlength{\abovecaptionskip}{0cm}
\caption{Results on multimodal concealed visual perception, where we conduct experiments on the RGB-Depth COD task. The simulated depth map is provided by PopNet \cite{wu2023source}. Following MultiCOS~\cite{fang2025integrating}, RUN~\cite{he2025run} and our RUN++ are adapted to process the multimodal inputs through simple structural modifications.}\label{table:DCODQuanti}
\setlength{\tabcolsep}{0.6mm}
\resizebox{\columnwidth}{!}{ 
\begin{tabular}{l|c|cccc|cccc|cccc|cccc} 
            \toprule
            \multicolumn{1}{c|}{} & \multicolumn{1}{c|}{} & \multicolumn{4}{c|}{\textit{CHAMELEON} }   & \multicolumn{4}{c|}{\textit{CAMO} }      & \multicolumn{4}{c|}{\textit{COD10K} }     & \multicolumn{4}{c}{\textit{NC4K} }         \\ 
            \cline{3-18} 
            \multicolumn{1}{l|}{\multirow{-2}{*}{Methods}}  & \multicolumn{1}{c|}{\multirow{-2}{*}{Pub.}}  &{\cellcolor{gray!40}$M$~$\downarrow$}                                  & {\cellcolor{gray!40}$F^{max}_\beta$~$\uparrow$}                               & {\cellcolor{gray!40}$E^{max}_\phi$~$\uparrow$}                               & \multicolumn{1}{c|}{\cellcolor{gray!40}$S_\alpha$~$\uparrow$}                  & {\cellcolor{gray!40}$M$~$\downarrow$}                                  & {\cellcolor{gray!40}$F^{max}_\beta$~$\uparrow$}                               & {\cellcolor{gray!40}$E^{max}_\phi$~$\uparrow$}                               & \multicolumn{1}{c|}{\cellcolor{gray!40}$S_\alpha$~$\uparrow$}                                   & {\cellcolor{gray!40}$M$~$\downarrow$}                                  & {\cellcolor{gray!40}$F^{max}_\beta$~$\uparrow$}                               & {\cellcolor{gray!40}$E^{max}_\phi$~$\uparrow$}                               & \multicolumn{1}{c|}{\cellcolor{gray!40}$S_\alpha$~$\uparrow$}                                   & {\cellcolor{gray!40}$M$~$\downarrow$}                                  & {\cellcolor{gray!40}$F^{max}_\beta$~$\uparrow$}                               & {\cellcolor{gray!40}$E^{max}_\phi$~$\uparrow$}                               & \multicolumn{1}{c}{\cellcolor{gray!40}$S_\alpha$~$\uparrow$}                                   \\ 
            \midrule 
            SPNet \cite{zhou2021specificity} & ICCV21 & 0.033 &0.872 &0.930 &0.888 &0.083 &0.807 &0.831 &0.783    &0.037 &0.776 &0.869 &0.808     &0.054 &0.828 &0.874 &0.825 \\ 
            SPSN \cite{lee2022spsn}  & ECCV22  & 0.032 &0.866 &0.932 &0.887      &0.084 &0.782 &0.829 &0.773    &0.042 &0.727 &0.854 &0.789     &0.059 &0.803 &0.867 &0.813 \\ 
            PopNet \cite{wu2023source} & ICCV23 & {0.022} & {0.893} & {0.962} & {0.910}  &0.073 &0.821 &0.869 &0.806    & {0.031} &0.789 &0.897 &0.827     &0.043 &0.852 &0.908 &0.852 \\ 
            DSAM \cite{yu2024exploring} & MM24 & 0.028 & 0.877 & 0.957 & 0.883  & {0.061} & {0.834} & {0.920} & {0.832}    &0.033 & {0.807} & {0.931} & {0.846}     & {0.040} & {0.862} & {0.940} & {0.871} \\              
            MultiCOS~\cite{fang2025integrating} & ArXiv25 &\color[HTML]{FF0000} \textbf{0.018} &\color[HTML]{00B0F0} \textbf{0.912} &{\color[HTML]{00B0F0} \textbf{0.970}} &{\color[HTML]{00B0F0} \textbf{0.923}}  & {{0.048}} &{{0.865}} &{{0.929}} &{{0.867}}   &{\color[HTML]{00B0F0} \textbf{0.020}} &{{0.850}} &{{0.946}} &{\color[HTML]{00B0F0} \textbf{0.880}} &{{0.031}} &{\color[HTML]{00B0F0} \textbf{0.882}} &{{0.944}} &{{0.890}}\\ 
            RUN~\cite{he2025run} & ICML25 & 0.019 & 0.908 & 0.967 & 0.922 &\color[HTML]{00B0F0} \textbf{0.047} &\color[HTML]{00B0F0} \textbf{0.872} &\color[HTML]{00B0F0} \textbf{0.933} &\color[HTML]{00B0F0} \textbf{0.869} &  0.021 &\color[HTML]{00B0F0} \textbf{0.853} & \color[HTML]{00B0F0} \textbf{0.947} & 0.879 &\color[HTML]{00B0F0} \textbf{0.030}  & 0.880 &\color[HTML]{00B0F0} \textbf{0.945} &\color[HTML]{00B0F0} \textbf{0.891}  \\
            \rowcolor{c2!20} RUN++ (Ours) & --- & \color[HTML]{FF0000} \textbf{0.018} & \color[HTML]{FF0000} \textbf{0.917} & \color[HTML]{FF0000} \textbf{0.973} & \color[HTML]{FF0000} \textbf{0.925} & \color[HTML]{FF0000} \textbf{0.045} & \color[HTML]{FF0000} \textbf{0.883} & \color[HTML]{FF0000} \textbf{0.939} & \color[HTML]{FF0000} \textbf{0.872} & \color[HTML]{FF0000} \textbf{0.018} & \color[HTML]{FF0000} \textbf{0.868} & \color[HTML]{FF0000} \textbf{0.957} & \color[HTML]{FF0000} \textbf{0.886} & \color[HTML]{FF0000} \textbf{0.028} & \color[HTML]{FF0000} \textbf{0.893} & \color[HTML]{FF0000} \textbf{0.957} & \color[HTML]{FF0000} \textbf{0.898}\\
            \bottomrule
    \end{tabular}}
\end{minipage}\vspace{-5mm}
\end{table*}
{\noindent \textbf{Transparent object detection}. 
TOD is critical, particularly for applications like autonomous driving where detecting transparent obstacles is essential for safe navigation. To evaluate our performance on this task, we conducted experiments on two datasets: \textit{GDD} and \textit{GSD}.
For a fair and direct comparison with contemporary methods, we adopted PVT V2 as the backbone. The training set was constructed using 2,980 images from \textit{GDD} and 3,202 images from \textit{GSD}, with the remaining images from both datasets forming the test set. Following the evaluation protocol established by GDNet~\cite{mei2020don}, we assessed performance using three metrics: $M$, mIoU, and maximum F-measure ($F_\beta^{max}$).
As demonstrated by the quantitative results in~\cref{table:TODQuanti} and the qualitative examples in~\cref{fig:MedQuali1}, our RUN++ surpasses existing methods on both datasets. The results show that we provide a more precise segmentation of transparent objects. This enhanced detection capability signifies a meaningful advancement with direct implications for improving the reliability and safety of autonomous driving systems.}

{\noindent \textbf{Concealed defect detection}. In this evaluation, we tested the model's ability to generalize from a related domain. Specifically, the model trained on the COD task was directly applied to segment defects in the \textit{CDS2K} dataset~\cite{fan2023advances} without any fine-tuning. Consistent with standard practice, we adopted PVT V2 as the default backbone. Performance was measured using six evaluation metrics, where higher values are better for all metrics except for MAE, for which a lower value indicates superior performance. As shown in ~\cref{table:CDDQuanti}, our method achieves superior results compared to existing state-of-the-art approaches and surpasses the second-best method (RUN) by $5.81\%$, which further validates the strong generalization performance of the RUN++ framework.}

\subsection{Comparisons in Label-Deficient CVP Tasks}
{CVP in label-deficient settings presents a significant challenge due to the scarcity of valid supervisory signals. Under these conditions, an effective segmenter must possess a robust capability in exploiting useful information from those limited labels. 
Concurrently, it must demonstrate resilience against the potentially corrupting influence of noisy pseudo-labels, which are often generated internally by semi-supervised or weakly-supervised frameworks. Hence, we also evaluate the generalization of our RUN++ in this task. In specifically, we integrate our RUN++ (also RUN~\cite{he2025run} for comparison) with the SEE framework~\cite{he2025segment} by replacing its own segmenter.

\begin{table*}[htp]
\begin{minipage}[c]{\textwidth}
\centering
\setlength{\abovecaptionskip}{0cm}
\caption{Results on video concealed visual perception. RUN~\cite{he2025run} and RUN++ follow the configuration of ZoomNext~\cite{pang2024zoomnext}.}\label{table:VCODQuanti}
\setlength{\tabcolsep}{1.3mm}
\resizebox{\columnwidth}{!}{ 
\begin{tabular}{l|c|cccccc|cccccc}
\toprule
                                              &                          & \multicolumn{6}{c|}{\textit{CAD}} & \multicolumn{6}{c}{\textit{MoCA-Mask-TE}} \\ \cline{3-14}
  \multirow{-2}{*}{Methods}                     & \multirow{-2}{*}{Pub.}   &\cellcolor{gray!40} S$_{m}~\uparrow$                  &\cellcolor{gray!40} F$^{\omega}_{\beta}~\uparrow$             &\cellcolor{gray!40} $M$$~\downarrow$ &\cellcolor{gray!40} $E^{max}_\phi~\uparrow$ & \cellcolor{gray!40} mDice$~\uparrow$ &\cellcolor{gray!40} mIoU$~\uparrow$ & \cellcolor{gray!40} S$_{m}~\uparrow$ & \cellcolor{gray!40} F$^{\omega}_{\beta}~\uparrow$ & \cellcolor{gray!40} $M$$~\downarrow$ &\cellcolor{gray!40} $E^{max}_\phi~\uparrow$ & \cellcolor{gray!40} mDice$~\uparrow$ & \cellcolor{gray!40} mIoU$~\uparrow$ \\
  \midrule
    PNS-Net~\cite{ji2021progressively} & MICCAI21     & 0.655                             & 0.325                                     & 0.048                    & 0.673            & 0.384            & 0.290           & 0.526            & 0.059                         & 0.035                    & 0.530            & 0.084            & 0.054           \\
  SINet-V2~\cite{fan2021concealed}              & TPAMI21   & 0.653                             & 0.382                                     & 0.039                     & 0.762            & 0.413            & 0.318           & 0.588            & 0.204                         & 0.031                & 0.642            & 0.245            & 0.180           \\
  MG~\cite{yang2021self}             & ICCV21               & 0.594                             & 0.336                                     & 0.059                    & 0.692            & 0.368            & 0.268           & 0.530            & 0.168                         & 0.067                & 0.561            & 0.181            & 0.127           \\
  STL-Net~\cite{cheng2022implicit}     & CVPR22   & 0.696                             & 0.481                                     & 0.030                      & {0.845}    & 0.493            & 0.402           & 0.631            & 0.311                         & 0.027               &\color[HTML]{00B0F0} {\textbf{0.759}}     & 0.360            & 0.272           \\
  ZoomNext~\cite{pang2024zoomnext}                           & TPAMI24         &{0.757}                     &{{0.593}}                              &\color[HTML]{FF0000} {\textbf{0.020}}          &  {{0.865}}     & {{0.599}}     &\color[HTML]{00B0F0}  {\textbf{0.510}}    &\color[HTML]{00B0F0}  {\textbf{0.734}}     & {{0.476}}                  &\color[HTML]{00B0F0}  {\textbf{0.010}}   & {{0.736}}   &\color[HTML]{00B0F0}  {\textbf{0.497}}            & {{0.422}}    \\
  RUN~\cite{he2025run} & ICML25 &\color[HTML]{00B0F0} \textbf{0.760} &\color[HTML]{00B0F0} \textbf{0.596} & 0.021 &\color[HTML]{00B0F0} \textbf{0.867} &\color[HTML]{00B0F0} \textbf{0.600} & 0.508 &  0.731 &\color[HTML]{00B0F0} \textbf{0.480} &{0.011} & 0.737 & \color[HTML]{00B0F0}  \textbf{0.516} &\color[HTML]{00B0F0} \textbf{0.423}    \\
\rowcolor{c2!20}   RUN++ (Ours) & --- &\color[HTML]{FF0000} \textbf{0.763} &\color[HTML]{FF0000} \textbf{0.602} &\color[HTML]{FF0000} \textbf{0.020} &\color[HTML]{FF0000} \textbf{0.883} &\color[HTML]{FF0000} \textbf{0.608} &\color[HTML]{FF0000} \textbf{0.522} 
&\color[HTML]{FF0000} \textbf{0.750} &\color[HTML]{FF0000} \textbf{0.530} &\color[HTML]{FF0000} \textbf{0.009} &\color[HTML]{FF0000} \textbf{0.801} &\color[HTML]{FF0000} \textbf{0.554}  &\color[HTML]{FF0000} \textbf{0.447} \\
  \bottomrule
\end{tabular}}
\end{minipage}\vspace{-3mm}
\end{table*}
\noindent \textbf{Weakly-supervised CVP}.
We conduct experiments on COD with scribble supervision and follow the setting of SEE~\cite{he2025segment}. We compare our RUN++ with cutting-edge algorithms and report the performance in \cref{table:WCODQuanti}. The results demonstrate that RUN++ outperforms all competing methods. Superior performance on \textit{COD10K} and \textit{CAMO}, which are drawn from the same datasets used in training, indicates the model's effectiveness in learning from sparse annotations. More importantly, the significant performance gains on the entirely unseen \textit{CHAMELEON} and \textit{NC4K} verify our strong generalization capability in open-set concealed object scenarios.

\noindent \textbf{Semi-supervised CVP}. In the semi-supervised setting, we also follow SEE~\cite{he2025segment} and draw similar conclusions. As shown in~\cref{table:WCODQuanti}, our RUN++ again achieves optimal results compared to other methods. This consistent, top-tier performance underscores the model's robustness and its advanced capacity to learn from incomplete data with deficient labels.}

\begin{table*}[t]
\begin{minipage}{.49\textwidth}
     \setlength{\abovecaptionskip}{0cm}
	\centering
	\includegraphics[width=\linewidth]{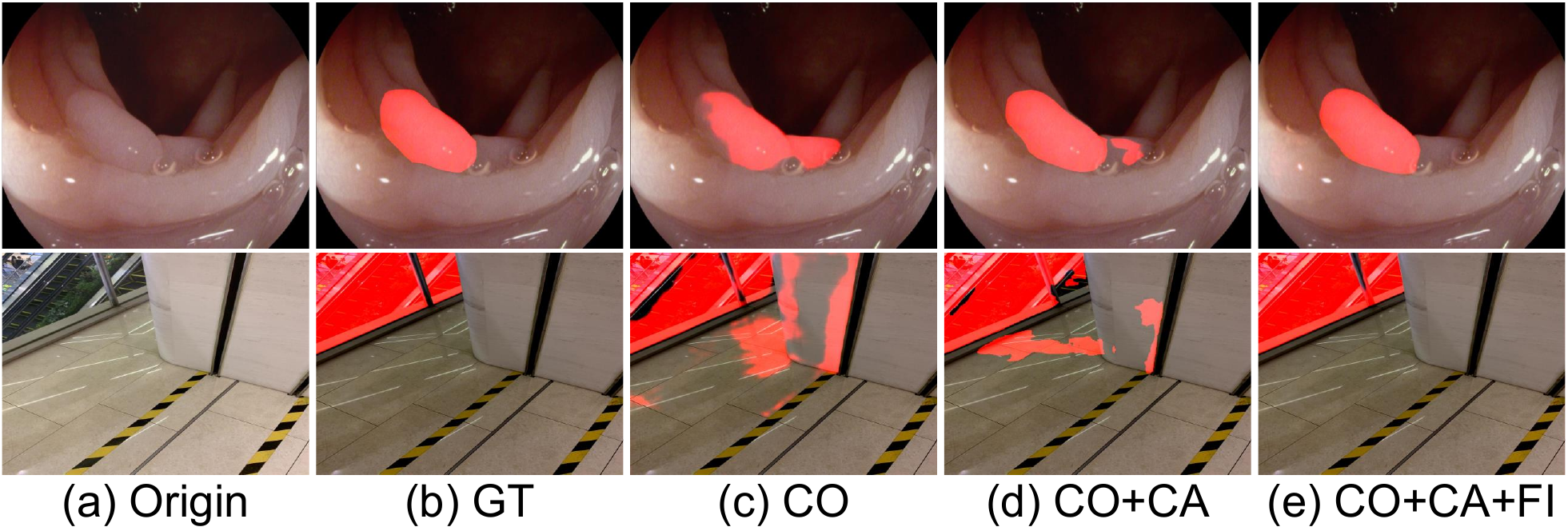}
	\captionof{figure}{Visualization of breakdown ablations, where CO, CA, and FI are short for CORE, CARE, and FINE, respectively. 
    (e) corresponds to our RUN++. 
    }
	\label{fig:BreakdownAblation}
	\vspace{1mm}   
    \end{minipage}
\begin{minipage}{.49\textwidth}
     \setlength{\abovecaptionskip}{0cm}
	\centering
	\includegraphics[width=\linewidth]{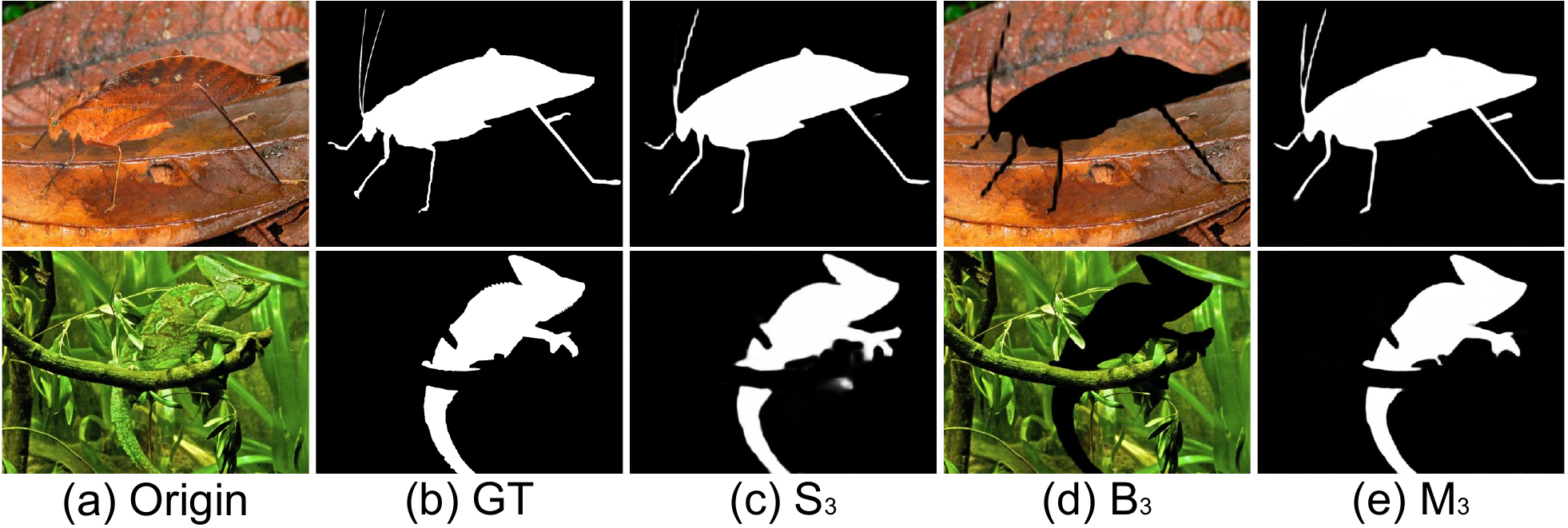}
	\captionof{figure}{Visualization of the segmentation pipeline at the final ($3^{rd}$) stage, where $\mathbf{S}_3$, $\mathbf{B}_3$, and $\mathbf{M}_3$ are the outputs from the CORE, CARE, and FINE modules and $\mathbf{M}_3$ is the final result. 
}
	\label{fig:VisualizationSeg}
	\vspace{1mm}   
    \end{minipage}
\\
\begin{minipage}[c]{0.345\textwidth}
\centering
		\setlength{\abovecaptionskip}{0cm}
		\caption{Breakdown ablations of RUN++.
		}
		\resizebox{\columnwidth}{!}{
			\setlength{\tabcolsep}{0.8mm}
\begin{tabular}{ccc|cccc}
\toprule
CORE & CARE & FINE & $M$~$\downarrow$  & $F_\beta$~$\uparrow$ & $E_\phi$~$\uparrow$ & $S_\alpha$~$\uparrow$   \\ \midrule
$\checkmark$    & $\times$    & $\times$    & 0.032 & 0.713   & 0.892   & 0.816 \\
$\times$    & $\checkmark$    & $\times$    & 0.038 & 0.683   & 0.855   & 0.797 \\
$\checkmark$    & $\checkmark$    & $\times$    & 0.030 & 0.745   & 0.902   & 0.828 \\
$\checkmark$    & $\checkmark$    & $\checkmark$    & \textbf{0.028} & \textbf{0.758}   & \textbf{0.913}   & \textbf{0.832} \\ \bottomrule
\end{tabular}}\label{table:BreakdownAblation}
\vspace{1mm}
\end{minipage}
\begin{minipage}[c]{0.652\textwidth}
\setlength{\abovecaptionskip}{0cm}
 \caption{Ablation study of the CORE module. }\label{Table:AblationCORE}
\resizebox{\columnwidth}{!}{ 
\setlength{\tabcolsep}{1.22mm}
\begin{tabular}{c|cccccc|c}
\toprule
Metrics & $\mathbf{I}\rightarrow E(\mathbf{I})$ & w/o RSS & w/o VSS & w/o prior $\hat{\mathbf{S}}_k$  & w/o prior $\mathbf{B}_{k-1}$  &\multicolumn{1}{c|}{w/o $\mathbf{E}_k$}  &    \cellcolor{c2!20} RUN++                  \\  \midrule
$M$~$\downarrow$     & 0.046                                    & 0.033                       & 0.030                       & 0.031                           & 0.030            &   0.029                &\cellcolor{c2!20} \textbf{0.028}                      \\
$F_\beta$~$\uparrow$         & 0.657                                    & 0.723                       & 0.746                       & 0.735                           & 0.747                          & 0.746                        & \cellcolor{c2!20} \textbf{0.758}                \\
$E_\phi$~$\uparrow$     & 0.845                                    & 0.893                       & 0.896                       & 0.889                           & 0.902                          & 0.901                        & \cellcolor{c2!20} \textbf{0.913}            \\
$S_\alpha$~$\uparrow$     & 0.760                                    & 0.812                       & 0.826                       & 0.823                           & 0.830                          & 0.825                        & \cellcolor{c2!20} \textbf{0.832}  \\ \bottomrule      
\end{tabular}} \label{table:AblationCORE}
\vspace{1mm}
\end{minipage}
\\
\begin{minipage}{0.4\textwidth}
\setlength{\abovecaptionskip}{0cm}
 \caption{Ablation study of the CARE module.}\label{Table:AblationCARE}
\resizebox{\columnwidth}{!}{ 
\setlength{\tabcolsep}{1mm}
\begin{tabular}{c|ccc|c}
\toprule
{Metrics} & $\mathcal{B}_1(\bigcdot) \rightarrow \mathcal{B}(\bigcdot)$ & $\mathcal{B}_2(\bigcdot) \rightarrow \mathcal{B}(\bigcdot)$ & \multicolumn{1}{c|}{w/o $\hat{\mathbf{I}}_k$} &    \cellcolor{c2!20} RUN++                  \\  \midrule
$M$~$\downarrow$     & \textbf{0.028}                 & \textbf{0.028}                 & 0.029         &\cellcolor{c2!20} \textbf{0.028}                       \\  
$F_\beta$~$\uparrow$         & 0.756                 & \textbf{0.760}                 & 0.749       & \cellcolor{c2!20} 0.758     \\ 
$E_\phi$~$\uparrow$     & 0.913                 & \textbf{0.915}                 & 0.903        & \cellcolor{c2!20}  0.913                       \\ 
$S_\alpha$~$\uparrow$     & \textbf{0.832}                 & 0.830                 & 0.827        &\cellcolor{c2!20}  \textbf{0.832}  \\ \bottomrule      
\end{tabular}} \vspace{1mm}
\end{minipage}
\begin{minipage}{0.588\textwidth}
\setlength{\abovecaptionskip}{0cm}
 \caption{Ablation study of the FINE module.}\label{Table:AblationFINE}
\resizebox{\columnwidth}{!}{ 
\setlength{\tabcolsep}{1mm}
\begin{tabular}{c|ccccc|c}
\toprule
{Metrics} &  w/o $\mathbf{U}_k$ & w/o $\mathbf{U}_k^{Inter}$ & w/o $\mathbf{U}_k^{Intra}$ &  $\mathbf{U}_k^1 \rightarrow \mathbf{U}_k$ & GDM$\rightarrow$BDM &    \cellcolor{c2!20} RUN++     \\  \midrule
$M$~$\downarrow$        & 0.031    & 0.029                           & 0.029                           & \textbf{0.028}                                      & 0.030                 &\cellcolor{c2!20} \textbf{0.028}                       \\  
$F_\beta$~$\uparrow$       & 0.737    & 0.745                           & 0.743                           & 0.750                                      & 0.742                 & \cellcolor{c2!20} \textbf{0.758}     \\ 
$E_\phi$~$\uparrow$       & 0.892    & 0.903                           & 0.901                           & 0.906                                      & 0.901                 & \cellcolor{c2!20}  \textbf{0.913}                       \\ 
$S_\alpha$~$\uparrow$      & 0.818    & 0.825                           & 0.826                           & 0.829                                      & 0.825                 &\cellcolor{c2!20}  \textbf{0.832}  \\ \bottomrule      
\end{tabular}} \vspace{1mm}
\end{minipage} \vspace{-5mm}
\end{table*}

\subsection{Comparisons in Multimodal and Video CVP Tasks}
{To further verify the generalization capabilities of our RUN++, we extended our evaluation to more complex visual perception tasks, including both multimodal and video concealed visual perception. These tasks introduce the challenge of processing multiple input streams—such as RGB-D data or sequential image frames—which contrasts with standard single-image CVP. To accommodate these multi-input scenarios, we adapted RUN++ (also RUN~\cite{he2025run}) with minimal structural modifications, adhering to established practices for each respective task, to demonstrate the inherent flexibility of our framework in handling more complex and dynamic data.

\noindent \textbf{Multimodal CVP}. For multimodal CVP, we conduct experiments on the RGB-Depth (RGB-D) COD task, utilizing simulated depth maps generated via PopNet~\cite{wu2023source}. 
To effectively leverage the supplementary depth information, we adapted the RUN++ architecture. Specifically, following the protocol of MultiCOS~\cite{fang2025integrating}, we employed a dual-encoder design using two PVT V2 backbones to extract features independently from the RGB image and its corresponding depth map. These features were subsequently fused using the attention module used in MultiCOS.
Following MultiCOS, performance was evaluated using four standard metrics: $M$, $F_\beta^{max}$, $E_\phi^{max}$, and $S_\alpha$. As detailed in~\cref{table:DCODQuanti}, RUN++ surpasses existing methods across all datasets, a result that underscores the architectural flexibility and strong adaptation capabilities of our framework.

\noindent \textbf{Video CVP}. To further demonstrate our versatility, we evaluate the effectiveness of RUN++ on video COD. Experiments are conducted on two standard benchmark datasets: \textit{CAD}~\cite{cheng2022implicit} and \textit{MoCA-Mask-TE}~\cite{bideau2016s}. 
Adhering to the experimental protocol of ZoomNext \cite{pang2024zoomnext}, we adopt PVT V2 as the feature extraction backbone and use 5-frame video clips as input. Performance is assessed using six standard metrics: $S_m$, $F^\omega_\beta$, $M$, $E^{max}_\phi$, mDice, and mIoU.
As depicted in~\cref{table:VCODQuanti}, our method achieves state-of-the-art performance, outperforming leading VCOS approaches. These results, along with the visualization results presented in~\cref{fig:MedQuali1}, not only confirm the flexibility of the RUN++ framework in adapting to dynamic video data but also highlight the superiority of our jointly reversible modeling strategy at both the mask and RGB levels.}

\begin{table*}
\begin{minipage}{.605\textwidth}
		\centering
		\setlength{\abovecaptionskip}{0cm}
		\caption{Effect of our CVP model (\cref{Eq:FinalModel}). CM, PM, OS, and DL are short for conventional model, proposed model, optimization solution, and deep learning (\textit{i.e.}, deep unfolding network), respectively.
		}
		\resizebox{\columnwidth}{!}{
			\setlength{\tabcolsep}{0.8mm}
			\begin{tabular}{c|ccccccc}
            \toprule
Metrics & PM+OS & CM1+DL & CM2+DL & CM3+DL & CM4+DL & CM5+DL & \cellcolor{c2!20}PM+DL (Ours) \\ \midrule
$M$~$\downarrow$   & 0.062 & 0.032  & 0.031  & 0.031  & 0.031  & 0.032  & \cellcolor{c2!20}\textbf{0.030}        \\
$F_\beta$~$\uparrow$ & 0.573 & 0.729  & 0.735  & 0.733  & 0.740  & 0.735  & \cellcolor{c2!20}\textbf{0.745}        \\
$E_\phi$~$\uparrow$   & 0.802 & 0.899  & 0.896  & 0.895  & 0.898  & 0.892  & \cellcolor{c2!20}\textbf{0.902}        \\
$S_\alpha$~$\uparrow$   & 0.733 & 0.823  & 0.824  & 0.821  & 0.823  & 0.823  & \cellcolor{c2!20}\textbf{0.828}  \\ \bottomrule
		\end{tabular}}\label{table:ablationModel}
		\vspace{1mm}
	\end{minipage} 
\begin{minipage}{.386\textwidth}
		    \centering
		\setlength{\abovecaptionskip}{0cm}
		\caption{Analysis of stage number $K$. We surpass most compared methods at $K=1$ and our performance improves as $K$ increases.
		}
		\resizebox{\columnwidth}{!}{
			\setlength{\tabcolsep}{0.6mm}
			\begin{tabular}{c|ccccc} \toprule
Metrics  & $K=1$    & \cellcolor{c2!20}$K=3$ (Ours) & $K=5$ & $K=7$ & $K=9$   \\  \midrule
$M$~$\downarrow$    & 0.030 &\cellcolor{c2!20} 0.028      & 0.027 & 0.027 & \textbf{0.026} \\
$F_\beta$~$\uparrow$  & 0.725 &\cellcolor{c2!20} 0.758      & 0.762 & 0.767 & \textbf{0.771} \\
$E_\phi$~$\uparrow$   & 0.893 &\cellcolor{c2!20} 0.913      & 0.913 & 0.916 & \textbf{0.920} \\
$S_\alpha$~$\uparrow$   & 0.822 &\cellcolor{c2!20} 0.832      & 0.835 & 0.836 & \textbf{0.838} \\ \bottomrule
		\end{tabular}}\label{table:ablationStageNum}
        \vspace{1mm}
		\end{minipage}\vspace{-5mm}
\end{table*}

\begin{table}
         \centering
		\setlength{\abovecaptionskip}{0cm}
		\caption{Analysis of iteration number $T$ of the Bernoulli diffusion model in our FINE module. 
		}
		\resizebox{\columnwidth}{!}{
			\setlength{\tabcolsep}{1mm}
			\begin{tabular}{c|cccccc} \toprule
Metrics  & $T=1$  & $T=3$   & \cellcolor{c2!20}$T=5$ (Ours) & $T=7$ & $T=9$ & $T=11$   \\  \midrule
$M$~$\downarrow$    & 0.030 & 0.029 &\cellcolor{c2!20} 0.028 & 0.028 & \textbf{0.027} & \textbf{0.027} \\
$F_\beta$~$\uparrow$  & 0.748 & 0.750 &\cellcolor{c2!20}  0.758 & 0.763 & 0.767 & \textbf{0.772} \\
$E_\phi$~$\uparrow$   & 0.904 & 0.907 &\cellcolor{c2!20}  0.913 & 0.915 & 0.917 & \textbf{0.921} \\
$S_\alpha$~$\uparrow$   & 0.828 & 0.830 &\cellcolor{c2!20}   0.832 & 0.833 & 0.835 & \textbf{0.836}\\ \bottomrule
		\end{tabular}}\label{table:ablationIterationNum}
        \vspace{-3mm}  
\end{table}

\begin{table}
\centering
		\setlength{\abovecaptionskip}{0cm}
		\caption{Results on small object images (1,084 images).
		}
		\resizebox{\columnwidth}{!}{
			\setlength{\tabcolsep}{1mm}
			\begin{tabular}{c|cccccc}
				\toprule
				Metrics & FEDER  & FGANet  & FocusDiff  &  FSEL& RUN & \cellcolor{c2!20} RUN++ (Ours)         \\ \midrule
				$M$~$\downarrow$  & 0.044 & 0.044  & 0.042     & 0.043 &{0.040} &\cellcolor{c2!20} \textbf{0.039} \\
				$F_\beta$~$\uparrow$ & 0.646 & 0.642  & 0.670     & 0.668 &{0.682} &\cellcolor{c2!20} \textbf{0.690} \\
				$E_\phi$~$\uparrow$ & 0.855 & 0.852  & 0.859     & 0.847 &{0.866} &\cellcolor{c2!20} \textbf{0.876} \\
				$S_\alpha$~$\uparrow$  & 0.777 & 0.776  & 0.781     & 0.776 &{0.789} &\cellcolor{c2!20} \textbf{0.792} \\ \bottomrule
		\end{tabular}}\label{table:small-object}
        \vspace{-1mm}  
\end{table}

\vspace{-2mm}
\subsection{Ablation Study}
We conduct ablation studies on \textit{COD10K} of the COD task.

{\noindent \textbf{Effect of basic network modules}. To validate the contributions of our key components, we conduct a series of breakdown studies. 
We first present a high-level analysis of the CORE, CARE, and FINE modules in~\cref{table:BreakdownAblation} and further provide corresponding visualizations of their impact on polyp and transparent object segmentation in~\cref{fig:BreakdownAblation}. 
A key qualitative finding, illustrated in~\cref{fig:VisualizationSeg}, is the contribution of the FINE module. While the segmentation mask $\mathbf{S}_3$ and the estimated background $\mathbf{B}_3$ can precisely capture the concealed object's general form, the final mask $\mathbf{M}_3$, after being refined by FINE, exhibits significantly finer details and sharper boundaries.
Furthermore, as presented in~\cref{Table:AblationCORE}, replacing deep features $E(\mathbf{I})$ with the raw input image leads to a significant performance decline, underscoring the importance of deep features. Furthermore, removing the RSS and VSS modules degrades performance, confirming the value of our state-space modeling. The benefit of our reversible strategy is also confirmed, as excluding either the foreground prior $\hat{\mathbf{S}}_k$ or the background prior $\mathbf{B}_{k-1}$ results in lower scores. Also, removing the auxiliary edge output brings performance decreases. 
Regarding the CARE module, as depicted in~\cref{Table:AblationCARE} we evaluate the choice of the reconstruction network. $\mathcal{B}(\bigcdot)$. Replacing it with more complex, large-scale networks (a CNN-based $\mathcal{B}_1(\bigcdot)$~\cite{he2023HQG} and a Transformer-based $\mathcal{B}_2(\bigcdot)$~\cite{he2023reti}) yields no significant performance gains, suggesting that a lightweight network is sufficient for preserving efficiency. Moreover, removing the reconstructed output $\hat{\mathbf{I}}_k$ entirely leads to suboptimal results, confirming the necessity of the image-level restoration process. 
Finally, for the FINE module, we validate both its uncertainty-guided strategy and kernel choice, and report the results in~\cref{Table:AblationFINE}. Replacing our entropy-based uncertainty map with a state-of-the-art estimation network~\cite{yang2021uncertainty} incurs extra computational overhead without improving performance, and removing the map entirely degrades results. We also confirm that a Bernoulli kernel suits the discrete segmentation refinement task better than a standard Gaussian kernel.}

{\noindent \textbf{Other configurations in RUN++}.  
We first validate the effect of our optimization model by comparing it against five alternative formulations (CM1-CM5), with quantitative results detailed in~\cref{table:ablationModel}. To isolate the impact of the optimization model, the FINE module was disabled for the experiments. CM1 corresponds to the baseline model in \cref{Eq:BasicModel1}. CM2 employs our full model from \cref{Eq:FinalModel} but replaces the proposed solver for the sparsity term $\mathcal{T}(\bigcdot)$ with a standard soft-thresholding method~\cite{he2023degradation}. 
CM3-CM5 represent ablations of the uncertainty-aware sparsity constraint: CM3 removes the uncertainty attention map $\mathbf{w}$; CM4 narrows the pixel value range for the filter from $[0.1,0.4) \cup (0.6,0.9]$ to $[0.1,0.3) \cup (0.7,0.9]$; and CM5 expands the filter range to all pixels. As demonstrated in~\cref{table:ablationModel}, our formulation consistently outperforms these alternative strategies, validating its superior design compared to both traditional solutions and other learning-based unfolding variants.\begin{table}[htbp!]
		    \centering
		\setlength{\abovecaptionskip}{0cm}
		\caption{Results on multi-object images (186 images).
		}
		\resizebox{\columnwidth}{!}{
			\setlength{\tabcolsep}{1mm}
			\begin{tabular}{c|cccccc}
				\toprule
				Metrics  & FEDER  & FGANet  &  FocusDiff  &  FSEL& RUN & \cellcolor{c2!20} RUN++ (Ours)     \\ \midrule
				$M$~$\downarrow$   & 0.068 & 0.065  & 0.062     & 0.062 &{0.060} &\cellcolor{c2!20} \textbf{0.059} \\
				$F_\beta$~$\uparrow$   & 0.480 & 0.481  & 0.500     & 0.496 &{0.505} &\cellcolor{c2!20} \textbf{0.513} \\
				$E_\phi$~$\uparrow$  & 0.813 & 0.810  & 0.818     & 0.820 &{0.827} &\cellcolor{c2!20} \textbf{0.832} \\
				$S_\alpha$~$\uparrow$    & 0.709 & 0.709  & 0.716     & 0.717 &{0.730} &\cellcolor{c2!20} \textbf{0.734} \\ \bottomrule
		\end{tabular}}\label{table:Multi-object}
        \vspace{-1mm}  
\end{table}
Moreover, as verified in~\cref{table:ablationStageNum,table:ablationIterationNum}, we analyze the optimal stage number for our RUN++ and the most suitable iterative diffusion timestep $T$ for our FINE module. To balance performance and computational efficiency, we set $K=3$ and $T=5$. 
}

\subsection{Further Analysis, Applications, and Meanings}

\noindent \textbf{Performance on small objects or multiple objects}. {Segmenting small objects and scenes with multiple objects presents a significant challenge in computer vision due to the inherent lack of discriminative visual cues~\cite{he2025segment}. To rigorously evaluate our method's performance under these specific conditions, we created two specialized test subsets by filtering the \textit{COD10K} dataset: \textit{(1) Small object subset}: This subset contains 1,084 images where the concealed object occupies less than $25\%$ of the total image area. \textit{(2) Multiple object subset}: This subset consists of 186 images that feature more than one concealed object. As detailed  in~\cref{table:small-object,table:Multi-object}, while a performance degradation was observed across all tested methods—an expected outcome given the increased difficulty—our approach consistently and significantly outperformed all competing methods. This demonstrates the superior robustness of our RUN++ in handling these challenging scenarios.
}

\begin{table*}[ht]
\centering
	\setlength{\abovecaptionskip}{0cm}
    \caption{Potential of RUN++ to serve as a mask refiner. SegR and SAMR are short for SegRefiner and SAMRefiner. FINE1 means using UGTR~\cite{yang2021uncertainty} for uncertainty estimation. Methods with the suffix ``+'' are retrained for the COD task. }
        \resizebox{\linewidth}{!}{
			\setlength{\tabcolsep}{1mm}
\begin{tabular}{l|cccccccccccc}
\toprule
Metrics & FEDER & + DCRF~\cite{krahenbuhl2011efficient} & + BS~\cite{barron2016fast}  & + SegR~\cite{wang2023segrefiner} & + SegR+~\cite{wang2023segrefiner} & + SAMR~\cite{lin2025samrefiner} & + SAMR+~\cite{lin2025samrefiner} & + RUN~\cite{he2025run}  & + FINE & + FINE1 &\cellcolor{c2!20} + RUN++  \\ \midrule
$M$~$\downarrow$ & 0.032 & 0.039 &  0.043 & 0.037      & 0.033        & 0.038      & 0.033        & 0.031 &  0.032 & 0.033  &\cellcolor{c2!20} 0.030 \\
$F_\beta$~$\uparrow$ & 0.715 & 0.683 &0.666  & 0.691      & 0.718        & 0.686      & 0.723        & 0.721 & 0.716 & 0.713  &\cellcolor{c2!20} 0.729  \\
$E_\phi$~$\uparrow$  & 0.892 & 0.853 & 0.862  & 0.863      & 0.889        & 0.857      & 0.906        & 0.897 & 0.896  & 0.897  &\cellcolor{c2!20} 0.910  \\
$S_\alpha$~$\uparrow$   & 0.810 & 0.797  &0.770 & 0.781      & 0.803        & 0.783      & 0.805        & 0.812 & 0.812   & 0.810 &\cellcolor{c2!20} 0.816 \\ \bottomrule
\end{tabular}}\vspace{-3mm}
\label{table:RefinerRUN}
\end{table*}

\begin{table*}[ht]
		\centering
		\setlength{\abovecaptionskip}{0cm}
		\caption{Generalization of RUN++. The suffixes``+'' and ``++'' means integrating the methods with RUN~\cite{he2025run} and our RUN++, respectively.
		}
		\resizebox{\linewidth}{!}{
			\setlength{\tabcolsep}{2mm}
			\begin{tabular}{l|ccc|ccc|ccc|ccc}
				\toprule
				Metrics &SINet & SINet+ &\cellcolor{c2!20} SINet++ & FEDER & FEDER+ &\cellcolor{c2!20} FEDER++ & FGANet & FGANet+ &\cellcolor{c2!20} FGANet++& FSEL & FSEL+ &\cellcolor{c2!20} FSEL++ \\ \midrule
				$M$~$\downarrow$ & 0.043 & 0.041 &\cellcolor{c2!20}0.040  &0.032 &0.031 &\cellcolor{c2!20} 0.030  & 0.032  &0.031 &\cellcolor{c2!20} 0.030  & 0.032 &  0.030 &\cellcolor{c2!20} 0.030 \\
				$F_\beta$~$\uparrow$ & 0.667 & 0.676 &\cellcolor{c2!20}0.687  &0.715 & 0.726 &\cellcolor{c2!20} 0.730 & 0.708  & 0.730 &\cellcolor{c2!20} 0.735  & 0.722 & 0.738 &\cellcolor{c2!20} 0.746  \\
				$E_\phi$~$\uparrow$ & 0.864 &0.870 &\cellcolor{c2!20}0.882  &0.892 & 0.902 &\cellcolor{c2!20} 0.910  & 0.894  & 0.901 &\cellcolor{c2!20} 0.912  & 0.891 & 0.905 &\cellcolor{c2!20} 0.915 \\
				$S_\alpha$~$\uparrow$ & 0.776 & 0.781 &\cellcolor{c2!20}0.788   &0.810 & 0.816 &\cellcolor{c2!20} 0.822  & 0.803  & 0.808 &\cellcolor{c2!20} 0.812  & 0.822 &\cellcolor{c2!20}0.830 &\cellcolor{c2!20} 0.832  \\ \bottomrule
		\end{tabular}}\label{table:GeneraRUN}\vspace{-2mm}
\end{table*}	

\begin{figure*}[ht]
	\centering
	\setlength{\abovecaptionskip}{0.1cm}
	\includegraphics[width=\linewidth]{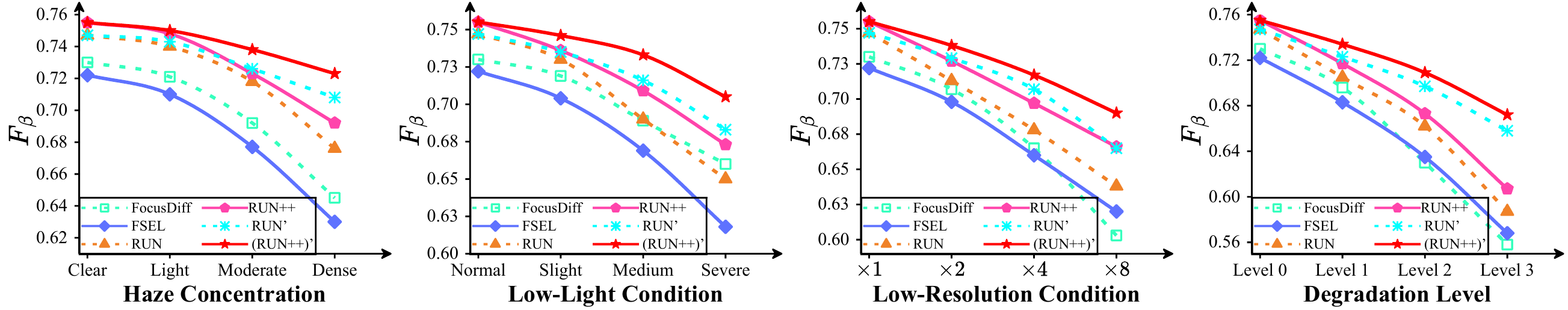}
	\caption{Performance on degraded CVP scenarios. RUN' and (RUN++)' are two degradation-resistant structures in~\cref{sec:BLO}.}\vspace{-5mm}
	\label{fig:DegradedCVP}
\end{figure*}

\begin{figure}[t]
     \setlength{\abovecaptionskip}{0cm}
	\centering
	\includegraphics[width=\linewidth]{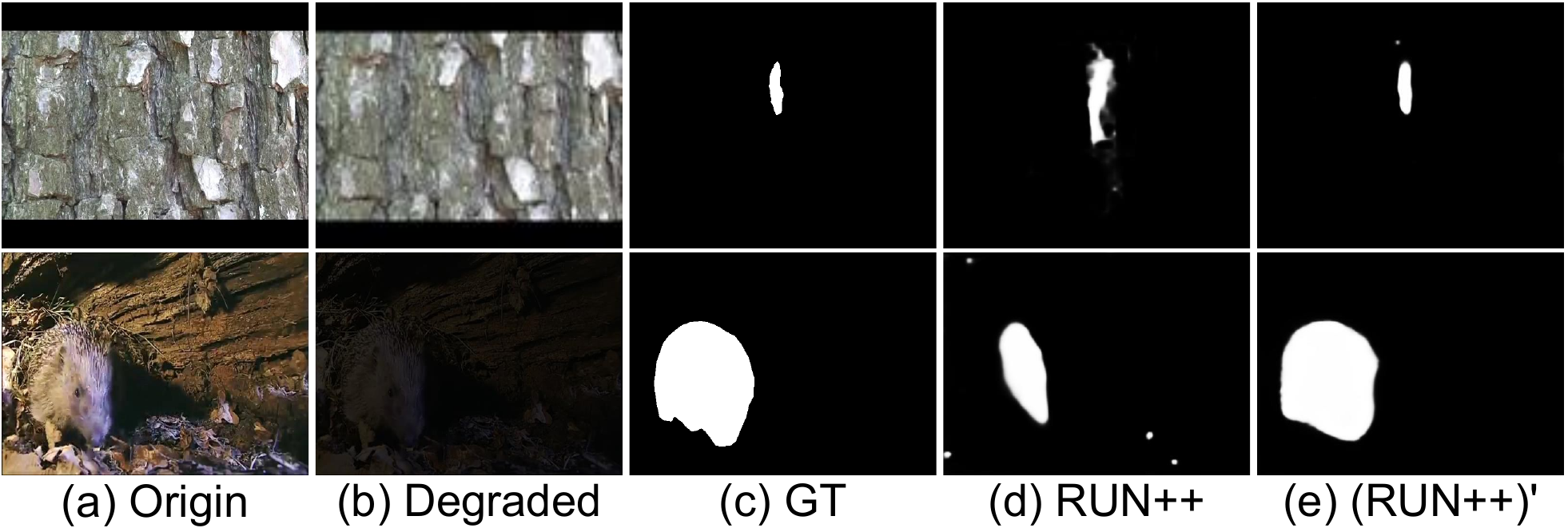}
	\captionof{figure}{Results on degraded CVP. First column: low-resolution (x4), Second column: low-light (Medium). 
    }
	\label{fig:DegradedCVPVis}
\vspace{-2mm}  
    \end{figure}

\noindent \textbf{Potential applications of RUN++}. 
{Here, we explore the versatility of the RUN++ framework by evaluating its performance in two key application scenarios: as a post-processing refiner for existing segmentation methods and as a plug-and-play architectural enhancement.
First, we assessed the efficacy of RUN++ as a refinement module. In this straightforward integration, we initialized the framework's input mask, $\mathbf{S}_0$, with the coarse segmentation output generated by an existing method (FEDER~\cite{He2023Camouflaged}). We compared this approach against both traditional refinement methods (dense crf (DCRF)~\cite{krahenbuhl2011efficient} and bilateral solver (BS)~\cite{barron2016fast}) and learning-based techniques (SegRefiner~\cite{wang2023segrefiner} and SAMRefiner~\cite{lin2025samrefiner}). For a fair comparison, and following original training protocols, we also present results for retrained versions of the learning-based refiners, denoted as SegRefiner+ and SAMRefiner+. 
As shown in~\cref{table:RefinerRUN}, the traditional methods and the pre-trained learning-based refiners failed to improve upon the initial coarse masks from FEDER. This outcome is largely attributable to the inherent difficulty of COD, a task characterized by a scarcity of discriminative cues. The initial coarse masks often suffer from significant quality issues, including mis-segmentations and blurred edges, which makes the refinement process exceptionally challenging.
After being retrained on the COD task, the learning-based refiners exhibited notable performance gains. However, while SAMRefiner+ achieved results comparable to RUN, it was still surpassed by our RUN++. Besides, only using the FINE module as a refiner with the uncertainty map calculated by entropy or estimated by a current method, UGTR~\cite{yang2021uncertainty}, cannot effectively refine the segmentation masks.
Next, we evaluated RUN++ as a foundational framework for enhancing existing models. This involved incorporating the core architectural components of other methods directly into the RUN++ unfolding structure and retraining the entire network from end to end.
As demonstrated in~\cref{table:GeneraRUN}, this deep integration yielded performance gains that significantly exceeded those of the simple refinement approach. Moreover, the improvements surpassed those achieved by the original RUN framework~\cite{he2025run}, demonstrating that our novel unfolding architecture can serve as an effective plug-and-play solution to substantially boost the performance of existing methods.}

\begin{table*}[t]
 \begin{minipage}[c]{\textwidth}
	\centering
	\setlength{\abovecaptionskip}{0cm}
	\caption{Restuls on salient object detection.  
    } \label{table:SODQuanti}
	\resizebox{\columnwidth}{!}{
		\setlength{\tabcolsep}{0.8mm}
		\begin{tabular}{l|cccc|cccc|cccc|cccc|cccc} 
			\toprule 
			\multicolumn{1}{l|}{}& \multicolumn{4}{c|}{\textit{DUT-OMRON} }& \multicolumn{4}{c|}{\textit{DUTS-test} }& \multicolumn{4}{c|}{\textit{ECSSD} }& \multicolumn{4}{c|}{\textit{HKU-IS} }& \multicolumn{4}{c}{\textit{PASCAL-S} }\\ \cline{2-21}
			\multicolumn{1}{l|}{\multirow{-2}{*}{Methods}} & {\cellcolor{gray!40}$M$~$\downarrow$} &{\cellcolor{gray!40}$F_\beta$~$\uparrow$} &{\cellcolor{gray!40}$E_\phi$~$\uparrow$} & \multicolumn{1}{c|}{\cellcolor{gray!40}$S_\alpha$~$\uparrow$}& {\cellcolor{gray!40}$M$~$\downarrow$} &{\cellcolor{gray!40}$F_\beta$~$\uparrow$} &{\cellcolor{gray!40}$E_\phi$~$\uparrow$} & \multicolumn{1}{c|}{\cellcolor{gray!40}$S_\alpha$~$\uparrow$}& {\cellcolor{gray!40}$M$~$\downarrow$} &{\cellcolor{gray!40}$F_\beta$~$\uparrow$} &{\cellcolor{gray!40}$E_\phi$~$\uparrow$} & \multicolumn{1}{c|}{\cellcolor{gray!40}$S_\alpha$~$\uparrow$}& {\cellcolor{gray!40}$M$~$\downarrow$} &{\cellcolor{gray!40}$F_\beta$~$\uparrow$} &{\cellcolor{gray!40}$E_\phi$~$\uparrow$} & \multicolumn{1}{c|}{\cellcolor{gray!40}$S_\alpha$~$\uparrow$}& {\cellcolor{gray!40}$M$~$\downarrow$} &{\cellcolor{gray!40}$F_\beta$~$\uparrow$} &{\cellcolor{gray!40}$E_\phi$~$\uparrow$} & \multicolumn{1}{c}{\cellcolor{gray!40}$S_\alpha$~$\uparrow$}\\ \midrule
VST~\cite{liu2021visual} & 0.058 & 0.755 & 0.871 &0.850 & 0.037 &0.828&0.919 & 0.896 & 0.033 & 0.910 & 0.951 & 0.932 & 0.029 & 0.897 & 0.952 & 0.928 & 0.061 & 0.816 & 0.902 & 0.872  \\
ICON-P~\cite{zhuge2022salient} & 0.047 & {\color[HTML]{00B0F0} \textbf{0.793}} & {\color[HTML]{00B0F0} \textbf{0.896}} & {{0.865}} & {\color[HTML]{00B0F0} \textbf{0.022}} & {{0.882}} & {{0.950}} & {\color[HTML]{00B0F0} \textbf{0.917}} & {{0.024}} & {{0.933}} & 0.964&0.940 & {\color[HTML]{00B0F0} \textbf{0.022}} & {{0.925}} & 0.967 & {{0.935}} & {\color[HTML]{00B0F0} \textbf{0.051}} & {\color[HTML]{FF0000} \textbf{0.847}} & {{0.921}} &  {{0.882}} \\
PGNet~\cite{xie2022pyramid} & {\color[HTML]{00B0F0} \textbf{0.045}} & 0.767 & 0.887 & 0.855 & 0.027 & 0.851 & 0.922 & 0.911 & {{0.024}} & 0.920 & 0.955 & 0.932 & 0.024 & 0.912 & 0.958 & 0.934 & 0.052 & 0.838 & 0.912 & 0.875 \\
MENet~\cite{wang2023pixels} & {\color[HTML]{00B0F0} \textbf{0.045}} & 0.782 & 0.891 & 0.849 & 0.028 & 
0.860 & 0.937 & 0.905& 0.033 & 0.906 & 0.954 & 0.928 &0.023 & 0.910 & 0.966 & 0.927 & 0.054 & 0.838 & 0.913 & 0.872\\
RMFormer~\cite{deng2023recurrent} & 0.049 & 0.775 & 0.892 & 0.862 & 0.030 & 0.850 & 0.928 & 0.907 & 0.028 & 0.917 & 0.957 & 0.933 & 0.024 & 0.908 & 0.960 & 0.930 &0.057 & 0.827 & 0.909 & 0.869 \\
GPONet~\cite{yi2024gponet} & {\color[HTML]{00B0F0} \textbf{0.045}} & 0.788 & 0.889 & {{0.865}} & 0.027 & 0.858 & 0.937 & 0.912 & 0.025 & 0.925 & 0.964 & {\color[HTML]{00B0F0} \textbf{0.942}} & 0.023 & 0.918 & 0.962 & {\color[HTML]{00B0F0} \textbf{0.936}} & 0.055 & 0.836 & 0.908 & 0.870   \\
VST-T++~\cite{liu2024vst++} &0.046&0.778& 0.892 & 0.853 &0.028 & 0.869& 0.943 & 0.901 &0.025&0.930& {{0.968}} & 0.937 & 0.024 & 0.919 & {{0.968}} & 0.930 & {\color[HTML]{00B0F0} \textbf{0.051}} & 0.841 & 0.901 & 0.878  \\
RUN~\cite{he2025run}& {\color[HTML]{00B0F0} \textbf{0.045}}  & 
 {\color[HTML]{00B0F0} \textbf{0.793}} &{{0.893}}  & {\color[HTML]{00B0F0} \textbf{0.867}}  & {\color[HTML]{00B0F0} \textbf{0.022}} & {\color[HTML]{00B0F0} \textbf{0.886}} & {\color[HTML]{00B0F0} \textbf{0.953}} & {{0.916}} & {\color[HTML]{00B0F0} \textbf{0.023}} & {\color[HTML]{00B0F0} \textbf{0.935}} & {\color[HTML]{00B0F0} \textbf{0.971}} & {{0.941}} & {\color[HTML]{00B0F0} \textbf{0.022}} & {\color[HTML]{00B0F0} \textbf{0.927}} & {\color[HTML]{00B0F0} \textbf{0.970}} & 0.934 & {\color[HTML]{00B0F0} \textbf{0.051}} & {{0.843}} & {\color[HTML]{00B0F0} \textbf{0.925}}  & {\color[HTML]{00B0F0} \textbf{0.882}} \\
 \rowcolor{c2!20}RUN++ (Ours) &\color[HTML]{FF0000} \textbf{0.044} &\color[HTML]{FF0000} \textbf{0.799} &\color[HTML]{FF0000} \textbf{0.901} &\color[HTML]{FF0000} \textbf{0.871} &\color[HTML]{FF0000} \textbf{0.021} &\color[HTML]{FF0000} \textbf{0.890} &\color[HTML]{FF0000} \textbf{0.957} &\color[HTML]{FF0000} \textbf{0.919} &\color[HTML]{FF0000} \textbf{0.022} &\color[HTML]{FF0000} \textbf{0.939} &\color[HTML]{FF0000} \textbf{0.976} &\color[HTML]{FF0000} \textbf{0.943} &\color[HTML]{FF0000} \textbf{0.021} &\color[HTML]{FF0000} \textbf{0.930} &\color[HTML]{FF0000} \textbf{0.975} &\color[HTML]{FF0000} \textbf{0.938} &\color[HTML]{FF0000} \textbf{0.050} &\color[HTML]{FF0000} \textbf{0.847} &\color[HTML]{FF0000} \textbf{0.927} &\color[HTML]{FF0000} \textbf{0.883} \\
   \bottomrule 
	\end{tabular}}
	\vspace{-0.5cm}
  \end{minipage}
\end{table*}
{\noindent \textbf{Performance on degraded CVP scenarios}. 
To evaluate performance under environmental degradation, we simulated four challenging scenarios on the \textit{COD10K} dataset, following established practices~\cite{he2023degradation,he2023reti,he2025unfoldir}. These scenarios include three specific corruptions—haze, low light, and low resolution—as well as a more challenging blind degradation setting, which applies a random combination of one to three of the preceding types.
We then assess the resilience of the compared methods under these adverse conditions.
As shown in Fig.~\ref{fig:DegradedCVP}, the performance of all methods declines as degradation intensity increases. Although our RUN++ demonstrates superior resilience—a result we attribute to its multi-modal reversible modeling and diffusion-based refinement—the framework still exhibits a sharp performance decrease under severe degradation. 
To enhance robustness, we employ the degradation-resistant framework detailed in~\cref{sec:BLO}. This involves replacing the reconstruction network, $\mathcal{B}(\bigcdot)$, with state-of-the-art and efficient restoration algorithms for specific conditions: CORUN~\cite{fang2024real} for haze, UnfoldIR~\cite{he2025unfoldir} for low light, and DiffIR~\cite{xia2023diffir} for low-resolution and blind degradation scenarios. Applying this to RUN~\cite{he2025run} and our RUN++ yields two fortified frameworks, which we denote as RUN' and (RUN++)', respectively.
Furthermore, the qualitative visualizations in Fig.~\ref{fig:DegradedCVPVis} corroborate these findings, demonstrating that our degradation-resistant framework produces markedly more accurate segmentation results. Together, these results underscore the effectiveness and potential of our degradation-resistant architecture for handling challenging real-world visual scenarios. }

\begin{figure}[t]
\setlength{\abovecaptionskip}{0cm}
	\centering
	\includegraphics[width=\linewidth]{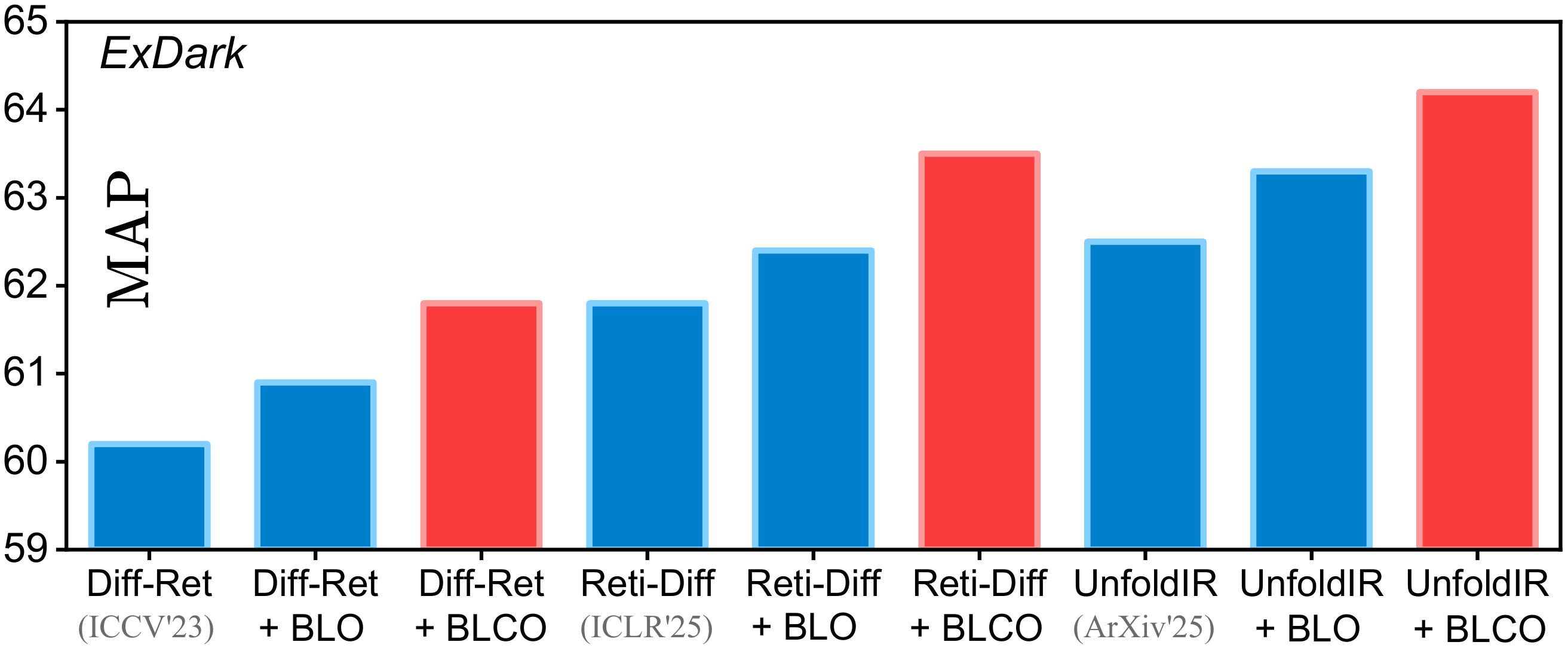}
	\caption{Application of the BLCO framework to low-light object detection, where BLO is a baseline that follows the practice from HQG-Net~\cite{he2023HQG} to construct.}
	\label{fig:lowlight}
	\vspace{-1mm}
\end{figure}

\begin{figure}[t]
     \setlength{\abovecaptionskip}{0cm}
	\centering
	\includegraphics[width=\linewidth]{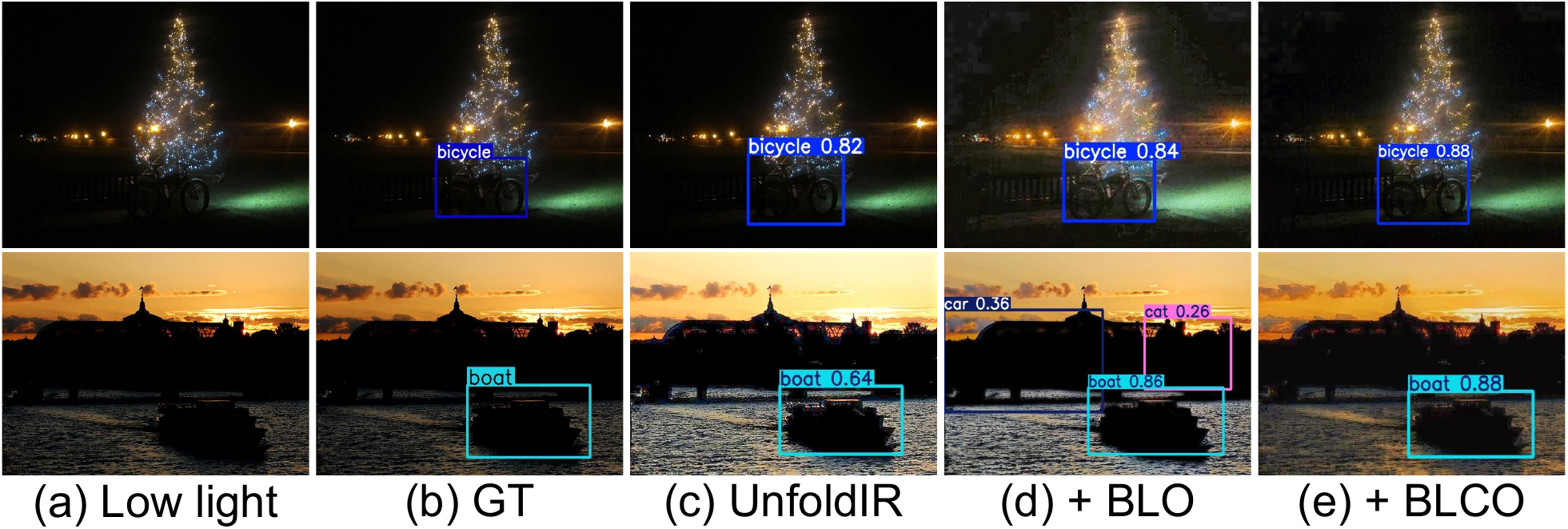}
	\captionof{figure}{Results on low-light object detection. Our BLCO framework achieves the best detection performance.
}
	\label{fig:VisualizationLow-light}
\vspace{-2mm}  
    \end{figure}

{\noindent \textbf{Application of BLCO to Low-Light Object Detection}. To demonstrate the generalizability of our BLCO principle, we extend it to low-light object detection. Our evaluation is conducted on the challenging \textit{ExDark} dataset~\cite{loh2019getting}, using the widely-adopted YOLO model as the baseline detector. We compare three distinct strategies for integrating state-of-the-art low-light enhancement methods—Diff-Retinex~\cite{yi2023diff}, Reti-Diff~\cite{he2023reti}, and UnfoldIR~\cite{he2025unfoldir}—with the detector. The first strategy is a tandem baseline, where the enhanced image is simply fed into the detector without any feedback. The second is the existing BLO framework introduced in UnfoldIR~\cite{he2025unfoldir} (the details can also be seen in~\cref{fig:discussion} (a)). The third is our proposed BLCO framework, which enhances the standard BLO approach by enabling direct feature-level interaction between the enhancement and detection networks to facilitate collaborative learning (\cref{fig:discussion} (b)). 
The results confirm the superiority of our BLCO. The quantitative data, presented in Fig.~\ref{fig:lowlight}, demonstrates that our BLCO framework consistently achieves the highest detection accuracy across all tested enhancement methods. These findings are corroborated by the qualitative examples in Fig.~\ref{fig:VisualizationLow-light}, which visually confirm the enhanced detection performance. This consistent outperformance validates the effectiveness of our collaborative optimization strategy, which leverages feature-level interaction to achieve more robust detection in challenging low-light conditions.}

{\noindent \textbf{Generalization on salient object detection}. We further evaluate the generalization of our method in SOD. Following the protocol established by RUN~\cite{he2025run}, we evaluated our model's performance on five widely-used benchmark datasets: \textit{DUT-OMRON} \cite{yang2013saliency}, \textit{DUTS-test} \cite{wang2017learning}, \textit{ECSSD} \cite{yan2013hierarchical}, \textit{HKU-IS} \cite{li2015visual}, and \textit{PASCAL-S} \cite{li2014secrets}. The evaluation metrics were the same as those used for the COD task. The results, presented in~\cref{table:SODQuanti}, show that our method outperforms existing state-of-the-art approaches and achieves a leading position. This outcome not only highlights the superiority of our framework but also underscores the significant potential of unfolding-based architectures for diverse high-level vision tasks.}

\section{Discussions on deploying DUNs in CVP tasks}

In addition to proposing the high-performance RUN++ method, this paper thoroughly explores the intrinsic advantages of Deep Unfolding Networks (DUNs) and investigates their deployment for concealed visual perception. From our analysis, we derive two significant conclusions:

First, as verified in Table~\ref{table:ablationModel}, incorporating \textbf{explicit regularization terms} into DUN-based frameworks yields superior performance compared to relying solely on implicit, data-driven priors. These explicit terms instill task-specific knowledge into the model, reducing the burden to learn complex regularities from data. Consequently, this strategy can lead to more parameter-efficient models and mitigates the limitations of lightweight networks in capturing sophisticated priors.

Second, as depicted in Fig.~\ref{fig:DegradedCVP} and Fig.~\ref{fig:lowlight}, integrating DUNs with \textbf{state-of-the-art low-level vision algorithms} is a highly effective strategy for enhancing performance in degraded scenarios. This can be conceptualized as adding an auxiliary branch, similar to an edge reconstruction branch that is commonly used~\cite{He2023Camouflaged}. While the low-level restoration task may have a significant domain gap with the primary high-level task, we find that the distinct perspective it provides offers complementary information. This supplementary guidance ultimately leads to more robust and accurate results.

More broadly, this highlights the value of our proposed Bi-Level Collaborative Optimization (BLCO) framework. Its principles of synergistic, co-dependent optimization are not limited to DUNs and can inspire the design of other advanced architectures. The BLCO framework provides a generalizable blueprint for developing robust vision systems capable of withstanding severe and unknown real-world degradations.

\section{Limitations and Future Work}
{While our method demonstrates robust segmentation performance, the integration of the reconstruction and diffusion-based refinement modules introduces non-negligible computational and storage overhead. A key direction for future work is therefore the development of more lightweight architectures that can maintain high performance without compromising efficiency, particularly in challenging degradation scenarios.

Furthermore, we identify two promising avenues for future research. The first involves investigating more advanced strategies for combining segmentation networks with low-level vision algorithms to enhance resilience against unknown, or ``blind'' degradations. The second direction is to develop probability-oriented models as a complement to our DUN-based approach. Such models could extend our framework to a broader range of high-level applications, such as image classification or retrieval, which do not require dense pixel-level understanding and are thus unsuited to the current foreground-background separation paradigm.}

\section{Conclusions}
{This paper introduces RUN++, a novel deep unfolding network that, for the first time, synergizes a reversible modeling strategy with a stage-wise diffusion model to address the CVP problem. We formulated the CVP task as a foreground-background separation problem, unfolding its optimization into a multi-stage network composed of CORE, CARE, and FINE modules. The CORE and CARE modules enable reversible modeling in the mask and RGB domains, respectively, while the innovative FINE module leverages a Bernoulli diffusion process to meticulously refine segmentation masks in uncertain regions. This unique combination balances the interpretability of a model-based unfolding approach with the powerful generative capabilities of diffusion models, creating a highly effective and efficient framework. Extensive experiments verify the superiority of our RUN++.}

\balance
\bibliographystyle{IEEEtran}
\bibliography{RUN++}

{\fontsize{8pt}{\baselineskip}\selectfont \noindent\textbf{Chunming He} received the B.S. degree from Nanjing University of Posts and Telecommunications, China, and the M.E. degree from Tsinghua University, China. He is currently a Ph.D. candidate at Duke University, Durham, USA. His research interests include computer vision and biomedical image analysis.}

{\fontsize{8pt}{\baselineskip}\selectfont \noindent\textbf{Fengyang Xiao} received the B.S. degree from Nanjing University of Posts and Telecommunications, China, and the M.S. degree from Sun Yat-sen University, China. She is currently a Ph.D. student at Duke University, Durham, USA. Her research interests include differential equations and numerical solutions, image processing and computer vision.
}

{\fontsize{8pt}{\baselineskip}\selectfont \noindent\textbf{Rihan Zhang} received his B.S. degree from Anhui University of Finance and Economics, China, and the M.S. degree from Anhui University, China. He is currently a research assistant at Duke University, Durham, USA. His research interests include computer vision and biomedical image analysis.
}

{\fontsize{8pt}{\baselineskip}\selectfont \noindent\textbf{Chengyu Fang} received the B.S. degree in software engineering from Southwest University, Chongqing, China in 2024. Now, he is pursuing his M.S. degree at Shenzhen International Graduate School, Tsinghua University. His research interests include computer vision and image processing.}

{\fontsize{8pt}{\baselineskip}\selectfont \noindent\textbf{Deng-Ping Fan} (Senior Member, IEEE) is a full professor and department chair of Computer Science and Technology at Nankai University. Previously, he was the Team Lead at the IIAI. 
Dr. Fan received his Ph.D. from Nankai University in 2019 and was a postdoctoral researcher at the ETH Zurich. He has published numerous high-quality papers, with approximately 29,000 citations on Google Scholar and an H-index of 57. 
Some representative awards include a World Artificial Intelligence Conference Youth Outstanding Paper and two CVPR best paper nominations. Dr. Fan serves as an AE for IEEE TIP and as an area chair for CVPR/NeurIPs. 
He has been listed among the top $2\%$ of global scientists by Stanford University.
}

{\fontsize{8pt}{\baselineskip}\selectfont \noindent\textbf{Sina Farsiu} (Fellow, IEEE) received his PhD degree in electrical engineering from the University of California, Santa Cruz in 2005. He is now the Anderson-Rupp Professor of BME, with secondary appointments at the Departments of ECE and CS at Duke University. He has served as the Senior Area Editor of IEEE Transactions on Image Processing, Deputy Editor of the Biomedical Optics Express, and the Associate Editor of SIAM Journal on Imaging Sciences. 
He received 
the “Outstanding Member of the Editorial Board Award” from the IEEE Signal Processing Society in 2018. He is a Fellow of IEEE, Optica, SPIE, ARVO, and AIMBE.}
\end{document}